\def\year{2021}\relax
\documentclass[letterpaper]{article} 
\usepackage{aaai21}  
\usepackage{times}  
\usepackage{helvet} 
\usepackage{courier}  
\usepackage[hyphens]{url}  
\usepackage{graphicx} 
\usepackage{natbib}  
\usepackage{caption} 
\usepackage{booktabs}       
\usepackage{amsfonts}       
\usepackage{nicefrac}       
\usepackage{microtype}      
\usepackage{multirow, makecell}
\usepackage{xcolor}
\usepackage{xspace}
\usepackage{enumitem}
\usepackage{amsmath}
\usepackage{mathtools}

\usepackage{enumitem}
\usepackage{bm}

\newcommand{\Db}{\mathbf{D}}

\newcommand{\Fb}{\mathbf{F}}

\newcommand{\Mb}{\mathbf{M}}

\newcommand{\Xb}{\mathbf{X}}

\newcommand{\hb}{\mathbf{h}}

\newcommand{\model}{Vid-ODE\xspace}
\newcommand{\task}{continuous-time video generation\xspace}

\title{Vid-ODE: Continuous-Time Video Generation \\ with Neural Ordinary Differential Equation}

\author {
    Sunghyun Park\textsuperscript{\rm 1}\textsuperscript{*},
    Kangyeol Kim\textsuperscript{\rm 1}\textsuperscript{*},
    Junsoo Lee\textsuperscript{\rm 1},\\
    Jaegul Choo\textsuperscript{\rm 1},
    Joonseok Lee\textsuperscript{\rm 2},
    Sookyung Kim\textsuperscript{\rm 3},
    Edward Choi\textsuperscript{\rm 1}\\
}
\affiliations {
    \textsuperscript{\rm 1}KAIST \\
    \textsuperscript{\rm 2}Google Research \\
    \textsuperscript{\rm 3}Lawrence Livermore Nat'l Lab.\\
    psh01087@kaist.ac.kr,
    kangyeolk@kaist.ac.kr,
    junsoolee93@kaist.ac.kr,\\
    jchoo@kaist.ac.kr,
    joonseok@google.com,
    kim79@llnl.gov,
    edwardchoi@kaist.ac.kr
}

\newcommand\blfootnote[1]{%
  \begingroup
  \renewcommand\thefootnote{}\footnote{#1}%
  \addtocounter{footnote}{-1}%
  \endgroup
}

\begin{document}

\maketitle

\blfootnote{\textsuperscript{*} These authors contributed equally.}

\begin{abstract}
Video generation models often operate under the assumption of fixed frame rates, which leads to suboptimal performance when it comes to handling flexible frame rates (\textit{e.g.,} increasing the frame rate of the more dynamic portion of the video as well as handling missing video frames).
To resolve the restricted nature of existing video generation models' ability to handle arbitrary timesteps, we propose continuous-time video generation by combining neural ODE (\texttt{\model}) with pixel-level video processing techniques.
Using ODE-ConvGRU as an encoder, a convolutional version of the recently proposed neural ODE, which enables us to learn continuous-time dynamics, \model can learn the spatio-temporal dynamics of input videos of flexible frame rates.
The decoder integrates the learned dynamics function to synthesize video frames at any given timesteps, where the pixel-level composition technique is used to maintain the sharpness of individual frames.
With extensive experiments on four real-world video datasets, we verify that the proposed \model outperforms state-of-the-art approaches under various video generation settings, both within the trained time range (interpolation) and beyond the range (extrapolation).
To the best of our knowledge, \model is the first work successfully performing continuous-time video generation using real-world videos.
\end{abstract}

\section{Introduction}
Videos, the recording of the continuous flow of visual information, inevitably discretize the continuous time into a predefined, finite number of units, \textit{e.g.,} 30 or 60 frames-per-second (FPS).
This leads to the development of rather rigid video generation models assuming fixed time intervals, restricting the modeling of underlying video dynamics.
Therefore it is challenging for those models to accept irregularly sampled frames or generate frames at unseen timesteps.
For example, most video generation models do not allow users to adjust the framerate depending on the contents of the video (\textit{e.g.}, higher framerate for the more dynamic portion).
This limitation applies not only to extrapolation (\textit{i.e.}, generating future video frames), but also to interpolation; given a 1-FPS video between $t=0$ and $t=5$, most existing models cannot create video frames at $t=1.5$ or $t=3.8$.

\begin{figure}[t]
    \begin{center}
    \centerline{\includegraphics[width=\linewidth]{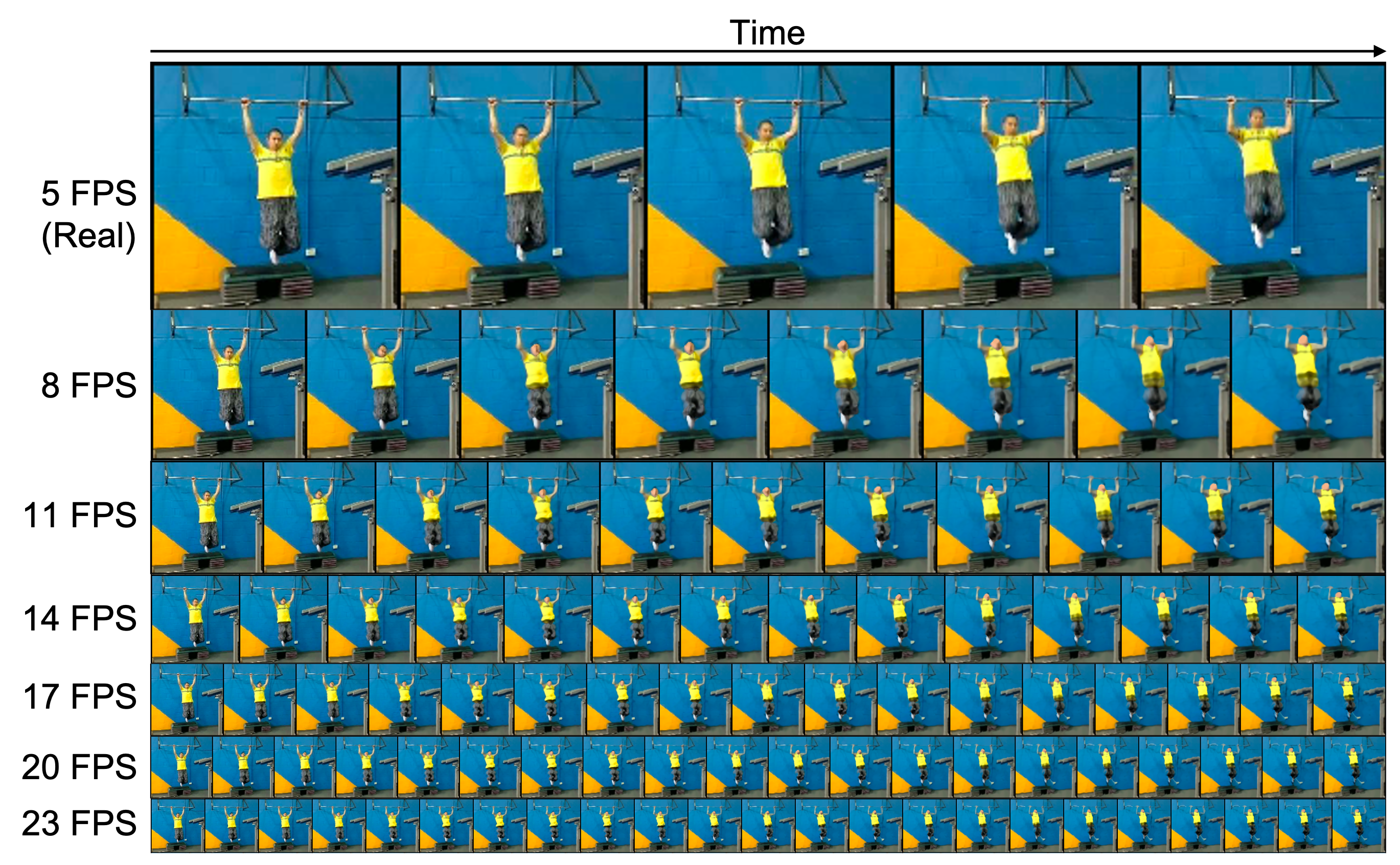}}
    \end{center}
    \label{fig:continuous interpolation}
    \caption{Generating video frames in diverse time intervals based on a 5 FPS video. (\emph{Top row:} Input to \model. \emph{Remaining rows:} Videos in various FPS between the start frame and the end frame.)}
\end{figure}

This might not seem like a serious limitation at first glance, since most videos we take and process are usually dense enough to capture the essential dynamics of actions.
However, video models are widely applied to understand spatio-temporal dynamics not just on visual recordings, but also on various scientific spatio-temporal data, which often do not follow the regular timestep assumption.

For instance, a climate video consists of multiple channels of climate variables (\textit{i.e.,} air pressure and ground temperature) instead of color density on the 2-D geographic grid. 
Due to the equipment cost, the time interval per each measurement often spans minutes to hours, which is insufficient to capture the target dynamics (\textit{e.g.,} creation and development of hurricanes). Consequently, existing video models often lead to sub-optimal predictions.
Another challenge with datasets collected from a wild environment is frequently missing values, which in turn results in irregular timesteps.

To resolve this limitation, we propose a video generation model based on Ordinary Differential Equation (\model) combined with a linear composition technique and adversarial training.
The proposed \model learns the continuous flow of videos from a sequence of frames (either regular or irregular) and is capable of synthesizing new frames at any given timesteps using the power of the recently proposed neural ODE framework, which handles the continuous flow of information~\cite{chen2018neural}.

Closely related to our work, ODE-RNN~\cite{rubanova2019latent} was recently proposed to handle arbitrary time gaps between observations, but limited to generating low-dimensional time-series data.
In order to predict high-dimensional spatio-temporal data, \model uses ODE convolutional GRU (ODE-ConvGRU), a convolution version of ODE-RNN, as an encoder to capture the spatio-temporal dynamics.
\model also employs adversarial training and a combination of pixel-level techniques such as optical flow and difference map to enhance the sharpness of the video.
Overall, \model is a versatile framework for performing \task with a single model architecture.

We summarize our contributions as follows:
\begin{itemize}[leftmargin=5.5mm]
    \item We propose \model that predicts video frames \emph{at any given timesteps} (both within and beyond the observed range).
    To the best of our knowledge, this is the first ODE-based framework to successfully perform \task on \emph{real-world videos}.
    \item According to extensive experiments on various real-world video datasets (\textit{e.g.,} human-action, animation, scientific data), \model consistently exhibits \emph{the state-of-the-art performance in \task}. With the ablation study, we validate the effectiveness of our proposed components along with their complementary roles.
    \item We demonstrate that \model can flexibly handle \emph{unrestricted by pre-defined time intervals} over the several variants of ConvGRU and neural ODEs on climate videos where data are sparsely collected.
\end{itemize}
\section{Related Work}

\textbf{Neural ordinary differential equations}
\model has parallels to neural ODE, an idea to interpret the forward pass of neural networks as solving an ODE, and several following works~\cite{rubanova2019latent, dupont2019augmented, de2019gru}. In particular, inspired by the application of neural ODE to the continuous time-series modeling, latent ODE~\cite{rubanova2019latent} equipped with ODE-RNN was proposed to handle irregularly-sampled time-series data. Recently, ODE$^{2}$VAE~\cite{yildiz2019ode2vae} attempted to decompose the latent representations into the position and the momentum to generate low-resolution image sequences. Although these prior works employing neural ODE show some promising directions in continuous time-series modeling, it is still an unanswered question whether they can scale to perform \task on complicated real-world videos, since existing methods demonstrated successful results only on small-scale synthetic or low-resolution datasets such as sinusoids, bouncing balls, or rotating MNIST. Our model aims at addressing this question by demonstrating the applicability in four real-world video datasets.

\noindent\textbf{Video Extrapolation}
The pixel-based video extrapolations, which are the most common approaches, predict each pixel from scratch, and often produce blur outcomes ~\cite{ballas2015delving, xingjian2015convolutional, lotter2016deep, wang2017predrnn, wang2018eidetic, wang2019memory, kwon2019predicting}.
Alternatively, motion-based methods~\cite{liu2017video, liang2017dual, gao2019disentangling}, which predict the transformation including an optical flow between two frames, generate sharp images, but the quality of outputs is degraded when it faces a large motion.
To tackle this, the models combining generated frames and transformed frames using linear composition~\cite{hao2018controllable} is proposed.

\noindent\textbf{Video Interpolation}
Conventional approaches~\cite{revaud2015epicflow} for interpolation often rely on hand-crafted methods such as a rule-based optical flow, resulting in limited applicability for real-world videos.
Recently, several neural-net-based approaches~\cite{dosovitskiy2015flownet, ilg2017flownet,jiang2018super,bao2019depth} exhibited a significant performance boost, taking advantage of end-to-end trainable models in a supervised fashion.
In addition, an unsupervised training method for video interpolation~\cite{reda2019unsupervised} was explored, providing an indirect way to train the neural networks.

\begin{figure*}[t!]
    \begin{center}
    \centerline{\includegraphics[width=\textwidth]{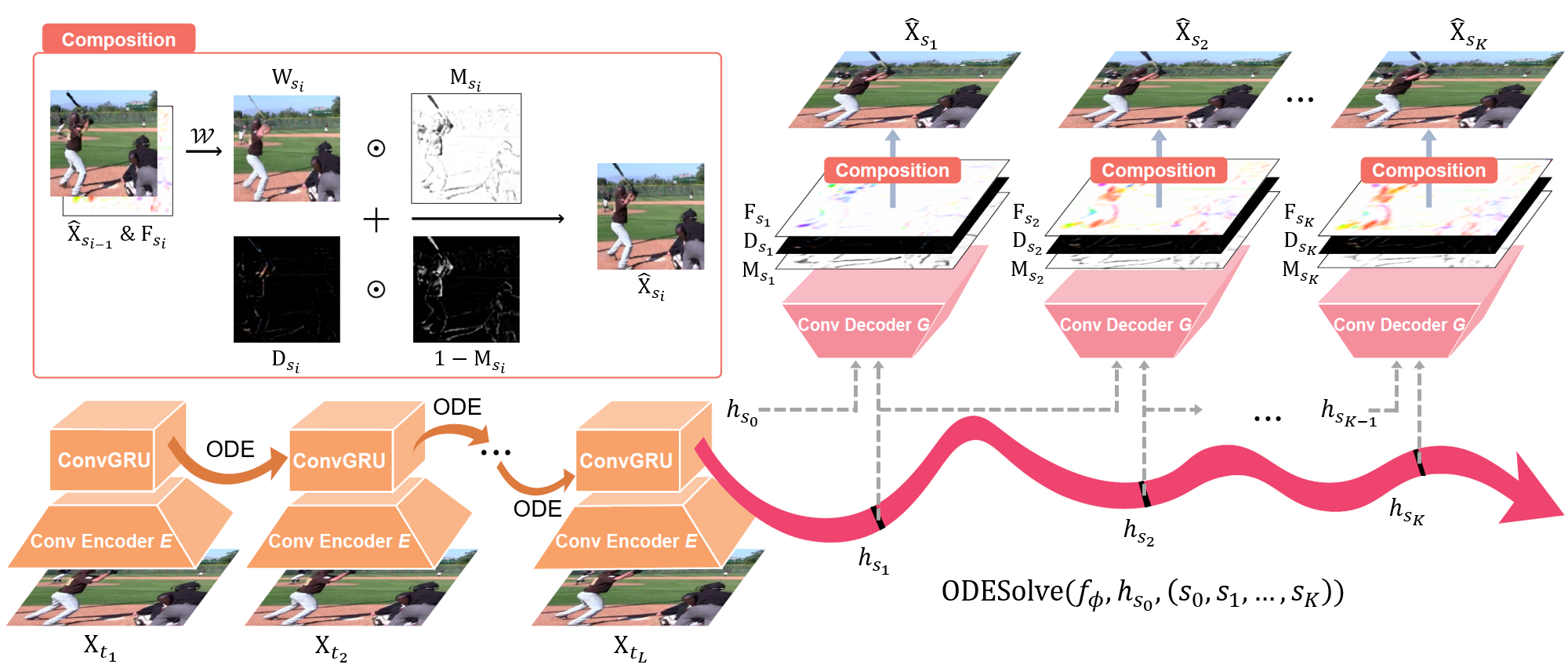}}
    \end{center}
    \caption{Overview of \model. First, input video frames $\mathcal{X}_\mathcal{T}$ are fed into a Conv-Encoder $E$, followed by ODE-ConvGRU.
    The final hidden state $\mathbf{h}_{T}$ is used as an initial value by another ODE solver, calculating the sequential hidden states $\mathbf{h}_{s_1}, \mathbf{h}_{s_2}, \ldots, \mathbf{h}_{s_K}$.
    Afterwards, the Conv-Decoder $G$ generates three intermediate representations $\mathbf{F}_{s_i}, \mathbf{D}_{s_i}, \mathbf{M}_{s_i}$ at each timestep $s_i$, which are combined via the linear composition $\Psi$ to generate target video frames $\{ \hat{\mathbf{X}}_{s_1}, \hat{\mathbf{X}}_{s_2}, \ldots, \hat{\mathbf{X}}_{s_K}\}$.}
    \label{fig:overview}
\end{figure*}

\section{Proposed Method: Video Generation ODE}

\textbf{Notations.} We denote $\mathcal{X}_\mathcal{T} \equiv \{\mathbf{X}_{t_1}, \mathbf{X}_{t_2}, ..., \mathbf{X}_{t_L}\}$ as a sequence of input video frames of length $L$, where each $\mathbf{X}_{i} \in \mathbb{R}^{m \times n \times c}$ is a 2-D image of size $m \times n$ with $c$ channels at irregularly sampled timesteps $\mathcal{T} \equiv \{ t_1, t_2, \dots, t_L \}$, where $0 < t_1 < t_2 < \dots < t_L$. We additionally define $t_0 = 0$ as origin, and specially denote the last timestep $t_L = T$.

\noindent\textbf{Problem Statement.}
Given an input video $\mathcal{X}_\mathcal{T}$, the goal of a continuous-time video generation problem is to generate video frames $\mathcal{X}_\mathcal{S}$ for another set of timesteps $\mathcal{S} \equiv \{ s_1, s_2, \dots, s_K \}$. As a couple of special cases, this task reduces to \emph{interpolation} if $0 \le s_i \le T$, but to  \emph{extrapolation} if $s_i > T$ for all $s_i \in \mathcal{S}$. 
Generally speaking, the query timesteps $\mathcal{S}$ may contain both inside and outside of the given range $\mathcal{T}$. 

\noindent\textbf{Overview of \model.}
As illustrated in Figure~\ref{fig:overview}, 
\model basically adopts an encoder-decoder structure.
First, the encoder embeds an input video sequence $\mathcal{X}_\mathcal{T}$ into the hidden state $\mathbf{h}_{T}$ using \emph{ODE-ConvGRU}, our novel combination of neural ODE~\cite{chen2018neural} and ConvGRU~\cite{ballas2015delving} (Section~\ref{sec:encoder}).
Then, from $\hb_T$, the decoder utilizes an \emph{ODE solver} to generate new video frames $ \hat{\mathbf{X}}_\mathcal{S}$ at arbitrary timesteps in $\mathcal{S}$ (Section~\ref{sec:decoder}).
Additionally, we include two discriminators in our framework to improve the quality of the outputs via  adversarial learning.
We end this section by describing our overall objective functions (Section~\ref{sec:loss}).

\subsection{Encoder: ODE-ConvGRU}
\label{sec:encoder}

Prior approaches elaborating neural ODE ~\cite{chen2018neural, rubanova2019latent, dupont2019augmented, de2019gru, yildiz2019ode2vae} employ a fully-connected network $f$ to model the derivative of the latent state $\mathbf{h}$ as 
\begin{equation}
  \frac{d\hb(t)}{dt} = f_{\theta}(\hb(t), t), \quad \hb(T) = \hb(0) + \int_{0}^{T} f_{\theta}(\hb(t), t) dt, \nonumber
\end{equation}
where $\theta$ is a set of trainable parameters of $f$.
Although this approach successfully models temporal dynamics in irregular timesteps, it is not capable of capturing the spatial dynamics of videos properly, making it difficult to produce a pixel-level reconstruction.
To tackle this issue, we propose the \emph{ODE-ConvGRU} architecture, a combination of neural ODE and ConvGRU specifically designed to capture not only temporal but also spatial information from the input videos. Formally, the ODE-ConvGRU is formulated as
\begin{align}
    \mathbf{h}_{t_i}^{-} &= \text{ODESolve}(f_{\theta}, \mathbf{h}_{t_{i-1}}, (t_{i-1}, t_{i})) \nonumber \\
    \mathbf{h}_{t_i} &= \text{ConvGRUCell}(\mathbf{h}_{t_i}^{-}, E(\mathbf{X}_{t_i})),
    \label{eq:encoder}
\end{align}
where $\mathbf{X}_{t_i} \in \mathbb{R}^{m \times n \times c}$ is an input video frame and $\mathbf{h}_{t_i} \in \mathbb{R}^{m' \times n' \times c'}$ is a latent state of size $m' \times n'$ with $c'$ channels
at $t_i$, for $i = 1, \dots, L$.
The ODE solver calculates the next hidden state $\mathbf{h}_{t_i}^{-}$ by integration based on $d\hb_{t_i}/dt$ which is approximated by a neural network $f_\theta$.
The initial hidden state $\mathbf{h}_0$ is set to zeros.
(Recall that $t_0 = 0$.)
Given an input frame $\mathbf{X}_{t_i}$, the Conv-Encoder $E$ produces an embedding $E(\mathbf{X}_{t_i}) \in \mathbb{R}^{m' \times n' \times c'}$ of this frame.
Taking this with the hidden state $\mathbf{h}_{t_i}^{-}$,
the ConvGRU cell derives the updated hidden state $\mathbf{h}_{t_i}$.
In our model, the RNN part of ODE-RNN~\cite{rubanova2019latent} is replaced with the ConvGRU cell to capture the spatio-temporal dynamics of video frames effectively.

\subsection{Decoder: ODE Solver + Linear Composition}
\label{sec:decoder}

The decoder generates a sequence of frames at target timesteps $\mathcal{S} \equiv \{ s_1, s_2, \dots, s_K \}$ based on the latent representation $\hb_{T}$ of the input video produced by the encoder. 
Our decoder consists of an \textit{ODE solver}, the \textit{Conv-Decoder} $G$, and the \textit{Linear composition} $\Psi$.

Formally, our decoder is described as
\begin{align}
    \mathbf{h}_{s_1}, \mathbf{h}_{s_2}, \ldots, \mathbf{h}_{s_K}  &= \text{ODESolve}(f_{\phi}, \mathbf{h}_{s_0}, (s_{1}, s_{2}, \ldots, s_K)), \nonumber \\
    \Fb_{s_i}, \Db_{s_i}, \Mb_{s_i} &= G(\mathbf{h}_{s_i}, \mathbf{h}_{s_{i-1}}), \nonumber \\
    \hat{\mathbf{X}}_{s_i} &= \Psi(\Fb_{s_i}, \Db_{s_i}, \Mb_{s_i}, \hat{\mathbf{X}}_{s_{i-1}}),
    \label{eq:decoder}
\end{align}
\begin{equation}
    \text{where } \Psi \coloneqq \underbrace{\mathbf{M}_{s_i}}_{\mathclap{\text{Composition Mask}}} \odot  \underbrace{\mathcal{W}(\overbrace{\mathbf{F}_{s_i}}^{\mathclap{\text{Optical Flow}}}, \hat{\mathbf{X}}_{s_{i-1}})}_{\text{Warped Image}} + (1-\mathbf{M}_{s_i}) \odot \underbrace{\mathbf{D}_{s_i}}_{\mathclap{\text{Image Difference}}}, \nonumber
\end{equation}
where $f_{\phi}$ is a convolutional neural network to approximate $d\hb_{s_i}/dt$ similarly to $f_{\theta}$ in the encoder. 
Given an initial value $\hb_{s_0} (=\hb_T)$, the ODE solver calculates the hidden representation $\hb_{s_i}$ at each timestep $s_{i}$, for $i=1,2,\ldots,K$.
Taking the current $\mathbf{h}_{s_i}$ and the previous $\mathbf{h}_{s_{i-1}}$, the Conv-Decoder $G$ produces three intermediate representations: optical flow $\Fb_{s_i}$, image difference $\Db_{s_i}$, and the composition mask $\Mb_{s_i}$. 
They are combined via the convex combination $\Psi$ to generate the final output frame $\hat{\mathbf{X}}_{s_i}$.
The details of $\Fb_{s_i}, \Db_{s_i}, \Mb_{s_i}$ and $\Psi$ are described as follows.

\noindent\textbf{Optical Flow} $(\Fb_{s_i})$. Optical flow is the vector field describing the apparent motion of each pixel between two adjacent frames. 
Compared to using static frames only, using the optical flow helps the model better understand the dynamics of the video to predict the immediate future or past by providing vector-wise information of moving objects~\cite{dosovitskiy2015flownet,ilg2017flownet,liu2017video}. 
Combined with a deterministic warping operation, optical flow helps the model preserve the sharpness of outputs.
In our model, we first predict the optical flow $\mathbf{F}_{s_i} \in \mathbb{R}^{m \times n \times 2}$ at an output timestep $s_i$.
Then, we apply the warping operation $\mathcal{W}$ on the previous generated image $\hat{\mathbf{X}}_{s_{i-1}}$, producing a warped image $\mathcal{W}(\mathbf{F}_{s_i}, \hat{\mathbf{X}}_{s_{i-1}})$.

\begin{figure}
    \centering
    \includegraphics[width=0.9\linewidth]{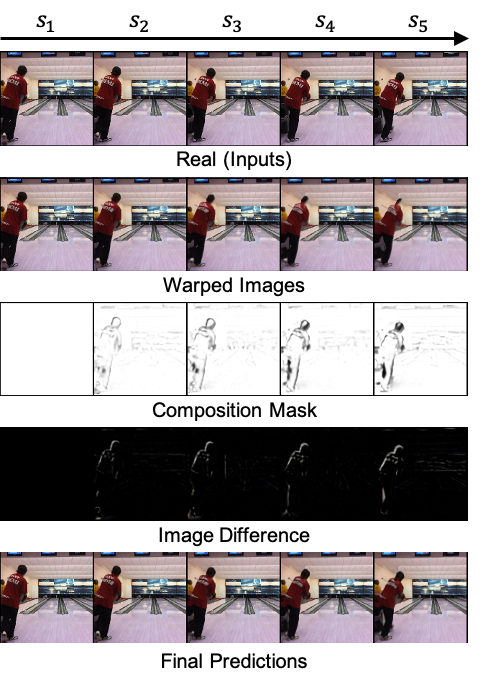}
    \caption{Visualization of three intermediate outputs (Warped Image, Image Difference, Composition Mask) during reconstructing images for video interpolation.}
    \label{fig:mask_visualization}
\end{figure}

\noindent\textbf{Image Difference} $(\Db_{s_i})$.
Although using optical flow helps to capture video dynamics, the quality of the warped image often degrades when it comes to generating completely new pixels or facing a large motion.
This is because, when there are big movements in the video, the warping operation moves a cluster of pixels to a new position and leaves an empty cluster behind.
This becomes even more problematic when we try to predict multiple frames in an autoregressive manner. 
To tackle this challenge, we employ the image difference $\mathbf{D}_{s_i} \in \mathbb{R}^{m \times n \times c}$ at each $s_i$, which predicts the pixel-wise difference between the current frame and the previous frame, $\Delta \Xb_{s_{i}}$.
Including $\mathbf{D}_{s_i}$ helps our model focus on areas where sudden movements take place.

\noindent\textbf{Composition Mask} $(\Mb_{s_i})$.
The final output frame $\hat{\Xb}_{s_i}$ is generated by combining the warped image $\mathcal{W}(\Fb_{s_i}, \hat{\Xb}_{s_{i-1}})$ and the image difference $\Db_{s_i}$ using element-wise convex combination weighted by the composition mask $\Mb_{s_i} \in \mathbb{R}^{m \times n}$.
While the other outputs of the Conv-Decoder $G$ (\textit{e.g.}, $\Fb_{s_i}, \Db_{s_i}$) have values in $[-\infty, \infty]$, $\Mb_{s_i}$ is forced to values in $[0,1]$ by a sigmoid activation, in order to perform its role as the convex combination weight.
Figure~\ref{fig:mask_visualization} depicts the role played by $\Fb_{s_i}, \Db_{s_i}$, and $\Mb_{s_i}$ when generating a video.
As seen in the warped images of Figure~\ref{fig:mask_visualization}, the moving portion (\textit{i.e.,} head) is gradually faded away as the image warping is repeatedly applied.
The image difference predicts the pixels, which can possibly have disappeared around large movements.
Through combining these two outputs, we can obtain the improved results as seen in the last row of Figure~\ref{fig:mask_visualization}.

\begin{table*}[t!]
    \centering
    \footnotesize
    \begin{tabular}{clcccccc}
    \toprule
    \multirow{2}{*}{\textbf{Datasets}}
    &\multicolumn{1}{c}{\multirow{2}{*}{\textbf{Model}}}&
    \multicolumn{3}{c}{\textbf{Video Interpolation}}&
    \multicolumn{3}{c}{\textbf{Video Extrapolation}}\\
    \cmidrule(lr){3-5} \cmidrule(lr){6-8}
     && SSIM$_{\uparrow}$ & LPIPS$_{\downarrow}$ & PSNR$_{\uparrow}$
     & SSIM$_{\uparrow}$ & LPIPS$_{\downarrow}$ & PSNR$_{\uparrow}$  \\
    \midrule
    \multirowcell{5}{\textit{KTH} \\ \textit{Action}}
    & Latent ODE
     & 0.730 &\ 0.481 &\ 20.99
     & 0.730 &\ 0.495 &\ 20.61
    \\
    & ODE$^2$VAE 
     & 0.752 &\ 0.456 &\ 23.28
     & 0.755 &\ 0.430 &\ 23.19
    \\
    & ODE-FC
     & 0.749 &\ 0.444 &\ 22.96
     & 0.750 &\ 0.442 &\ 23.00
    \\
    & ODE-Conv
     & 0.769 &\ 0.416 &\ 25.12
     & 0.768 &\ 0.429 &\ 24.31
    \\
    \cmidrule(lr){2-8}
    & \texttt{\model}
     & \textbf{0.911} &\ \textbf{0.048} &\ \textbf{31.77}
     & \textbf{0.878} &\ \textbf{0.080} &\ \textbf{28.19}
    \\
    \midrule
    \multirowcell{5}{\textit{Moving} \\ \textit{GIF}}
    & Latent ODE
     & 0.700 &\ 0.483 &\ 15.69
     & 0.675 &\ 0.513 &\ 14.41
    \\
    & ODE$^2$VAE
     & 0.715 &\ 0.456 &\ 16.49
     & 0.704 &\ 0.471 &\ 15.91
    \\
    & ODE-FC
     & 0.717 &\ 0.446 &\ 16.55
     & 0.704 &\ 0.452 &\ 15.86
    \\
    & ODE-Conv
     & 0.745 &\ 0.380 &\ 17.88
     & 0.713 &\ 0.429 &\ 16.23
    \\
    \cmidrule(lr){2-8}
    & \texttt{\model}
     & \textbf{0.815} &\ \textbf{0.115} &\ \textbf{18.44}
     & \textbf{0.778} &\ \textbf{0.156} &\ \textbf{16.68}
    \\
    \midrule
    \multirowcell{5}{\textit{Penn} \\ \textit{Action}}
    & Latent ODE
     & 0.377 &\ 0.762 &\ 15.63
     & 0.374 &\ 0.775 &\ 15.32
    \\
    & ODE$^2$VAE  
     & 0.433 &\ 0.687 &\ 17.06
     & 0.423 &\ 0.701 &\ 16.90
    \\
    & ODE-FC 
     & 0.447 &\ 0.643 &\ 17.40
     & 0.342 &\ 0.753 &\ 15.00
    \\
    & ODE-Conv
     & 0.550 &\ 0.538 &\ 19.25
     & 0.557 &\ 0.514 &\ 19.23
    \\
    \cmidrule(lr){2-8}
    & \texttt{\model}
     & \textbf{0.920} &\ \textbf{0.033} &\ \textbf{26.73}
     & \textbf{0.880} &\ \textbf{0.045} &\ \textbf{23.81}
    \\
    \bottomrule
    \end{tabular}
    \caption{Comparison with neural-ODE-based models}
    \label{Table:2:comparison_ode}
\end{table*}

\subsection{Objective Functions}
\label{sec:loss}

\noindent\textbf{Adversarial Loss}\footnote{Note that the adversarial loss formulation represented here is for video extrapolation. The formulation for video interpolation is provided in the supplementary material.}
\label{sec:gan}
We adopt two discriminators, one at the image level and the other at the video sequence level, to improve the output quality both in spatial appearance and temporal dynamics.
The image discriminator $D_\text{img}$ distinguishes the real image $\mathbf{X}_{s_i}$ from the generated image $\hat{\mathbf{X}}_{s_i}$ for each target timestep $s_i$. 
The sequence discriminator $D_{\text{seq}}$ distinguishes a real sequence $\mathcal{X}_\mathcal{S}$ from the generated sequence $\hat{\mathcal{X}}_\mathcal{S}$ for all timesteps in $\mathcal{S}$. 
Specifically, we adopt LS-GAN~\cite{mao2017least} to model $D_{\text{img}}$ and $D_{\text{seq}}$ as
{ \small
\begin{align}
    &\min_{\footnotesize \text{\model}} \max_{D_{\text{img}}}
    \mathcal{L}^{\text{img}}_{\text{adv}} = 
    \mathbb{E}_{\mathbf{X}_{s_i} \sim p(\mathcal{X}_\mathcal{S})}
    \left[ \left( D_{\text{img}} \left(\mathbf{X}_{s_i} \right) - 1 \right)^2 \right] \\ 
    & \hspace{2.0cm} + \mathbb{E}_{\mathcal{X}_\mathcal{T} \sim p(\mathcal{X}_\mathcal{T})}
    \left[ \left( D_{\text{img}}({\footnotesize \text{\model}}(\hat{\mathbf{X}}_{s_i}|\mathcal{X}_\mathcal{T})) \right)^2 \right] \nonumber \\
    &\min_{\footnotesize \text{\model}} \max_{D_{\text{seq}}}
    \mathcal{L}^{\text{seq}}_{\text{adv}} = 
    \mathbb{E}_{\mathcal{X}_{t_i:s_i} \sim p(\mathcal{X}_\mathcal{T;S})}
    \left[ \left( D_{\text{seq}}(\mathcal{X}_{t_i:s_i})-1 \right)^2 \right] \\
    &+ \mathbb{E}_{\mathcal{X}_{t_{i}:s_{i-1}} \sim p(\mathcal{X}_\mathcal{T;S})}
    \left[ \left( D_{\text{seq}} \big(\mathcal{X}_{t_i:s_{i-1}};{\footnotesize \text{\model}}(\hat{\mathbf{X}}_{s_i}|\mathcal{X}_\mathcal{T})\big) \right)^2 \right], \nonumber 
    \label{eq:adv_loss}
\end{align}}
where $\mathcal{T;S}$ is union of the timesteps $\mathcal{T}$ and $\mathcal{S}$, and $\mathcal{X}_{t_i:s_i}$ is a sequence of frames from $t_i$ to $T$, concatenated with frames from $s_1$ to $s_i$, for some $i = 1, \ldots, K$.

\noindent\textbf{Reconstruction Loss.}
$\mathcal{L}_\text{recon}$ computes the pixel-level $L_1$ distance between the predicted video frame $\hat{\mathbf{X}}_{s_i}$ and the ground-truth frame $\mathbf{X}_{s_i}$. Formally,
\begin{equation}
  \mathcal{L}_{\text{recon}} = \mathbb{E}_{\mathbf{X}_{s_i} \sim \mathcal{X}_\mathcal{S}}
  \left[ \| \hat{\mathbf{X}}_{s_i} - \mathbf{X}_{s_i} \|_{1} \right].
\end{equation}

\noindent\textbf{Difference Loss}    
$\mathcal{L}_\text{diff}$ helps the model learn the image difference $\mathbf{D}_{s_i}$ as the pixel-wise difference between consecutive video frames. Formally,
\begin{equation}
    \mathcal{L}_{\text{diff}} = 
    \mathbb{E}_{\Delta \mathbf{X}_{s_i} \sim \Delta\mathcal{X}_\mathcal{S}}
    \left[ \| \mathbf{D}_{s_i} - \Delta \mathbf{X}_{s_i} \|_{1} \right],
    \label{eq:diff_loss}
\end{equation}
where $\Delta \mathbf{X}_{s_i}$ denotes the image difference between two consecutive frames, \textit{i.e.,} $\mathbf{X}_{s_i} - \mathbf{X}_{s_{i-1}}$.

\noindent\textbf{Overall Objective}
\model is trained end-to-end using the following objective function:
\begin{equation}
    \mathcal{L} =
        \mathcal{L}_{\text{recon}} + 
        \lambda_{\text{diff}}\mathcal{L}_{\text{diff}} + 
        \lambda_{\text{img}}\mathcal{L}^{\text{img}}_{\text{adv}} +
        \lambda_{\text{seq}}\mathcal{L}^{\text{seq}}_{\text{adv}}, 
        \label{eq:objective}
\end{equation}
where we use $\lambda_{\text{diff}}$, $\lambda_{\text{img}}$, and $\lambda_{\text{seq}}$ for hyper-parameters controlling relative importance between different losses.
\section{Experiments}
\label{sec:exp}

\subsection{Experimental Setup}
We evaluate our model on two tasks: video interpolation and video extrapolation. 
In video interpolation, a sequence of five input frames are given, and the model is trained to reconstruct the input frames during the training phase. At inference, it predicts the four intermediate frames between the input time steps.
In video extrapolation, given a sequence of five input frames, the model is trained to output the next five future frames, which are available in training data. 
At inference, it predicts the next five frames.
We conclude this section with the analysis of the role of each component of \model, including an ablation study.

\noindent\textbf{Evaluation Metrics.} We evaluate our model using two metrics widely-used in video interpolation and video extrapolation, including Structural Similarity (SSIM)~\cite{wang2004image}, Peak Signal-to-Noise Ratio (PSNR).
In addition, we use Learned Perceptual Image Patch Similarity (LPIPS)~\cite{zhang2018unreasonable} to measure a semantic distance between a pair of the real and generated frames. 
Higher is better for SSIM and PSNR, lower is better for LPIPS.

\noindent\textbf{Datasets.} For our evaluation, we employ and preprocess the four real-world datasets and the one synthetic dataset as follows:

\textbf{KTH Action}~\cite{schuldt2004} consists of 399 videos of 25 subjects performing six different types of actions (walking, jogging, running, boxing, hand waving, and hand clapping).
We use 255 videos of 16 (out of 25) subjects for training and the rest for testing.
The spatial resolution of this dataset is originally $160 \times 120$, but we center-crop and resize it to $128 \times 128$ for both training and testing.

\textbf{Moving GIF}~\cite{siarohin2019animating} consists of 1,000 videos of animated animal characters, such as tiger, dog, and horse, running or walking in a white background. We use 900 for training and 100 for testing.
The spatial resolution of the original dataset is $256 \times 256$, and each frame is resized to $128 \times 128$ pixels.
Compared to other datasets, Moving GIF contains relatively larger movement, especially in the legs of cartoon characters. 

\textbf{Penn Action}~\cite{zhang2013} consists of videos of humans playing sports. The dataset contains 2,326 videos in total, involving 15 different sports actions, including baseball swing, bench press, and bowling. The resolution of the frames is within the size of $640 \times 480$. For training and testing, we center-crop each frame and then resize it to $128 \times 128$ pixels. We use 1,258 videos for training and 1,068 for testing.

\textbf{CAM5}~\cite{kim2019} is a hurricane video dataset, where we evaluate our model on video extrapolation for irregular-sampled input videos.
This dataset contains the frames of the global atmospheric states for every 3 hours with around $0.25^\circ$ resolution, using the annotated hurricane records from 1996 to 2015. 
We use zonal wind (U850), meridional wind (V850), and sea-level pressure (PSL) out of multiple physical variables available in each frame.
We take only those time periods during which hurricane actually occurs, resulting in 319 videos.
We use 280 out of these for training and 39 for testing. 
To fit large-scale global climate videos into GPU memory, we split the global map into several non-overlapping basins of $60^\circ \times 160^\circ$ sub-images.

\textbf{Bouncing Ball} contains three balls moving in different directions with a resolution of $32 \times 32$, where we evaluate our model for video extrapolation when non-linear motions occur as videos proceed. 
We use 1,000 videos for training and 50 videos for testing.

\begin{figure*}[t!]
    \centering
    \centerline{\includegraphics[width=\linewidth]{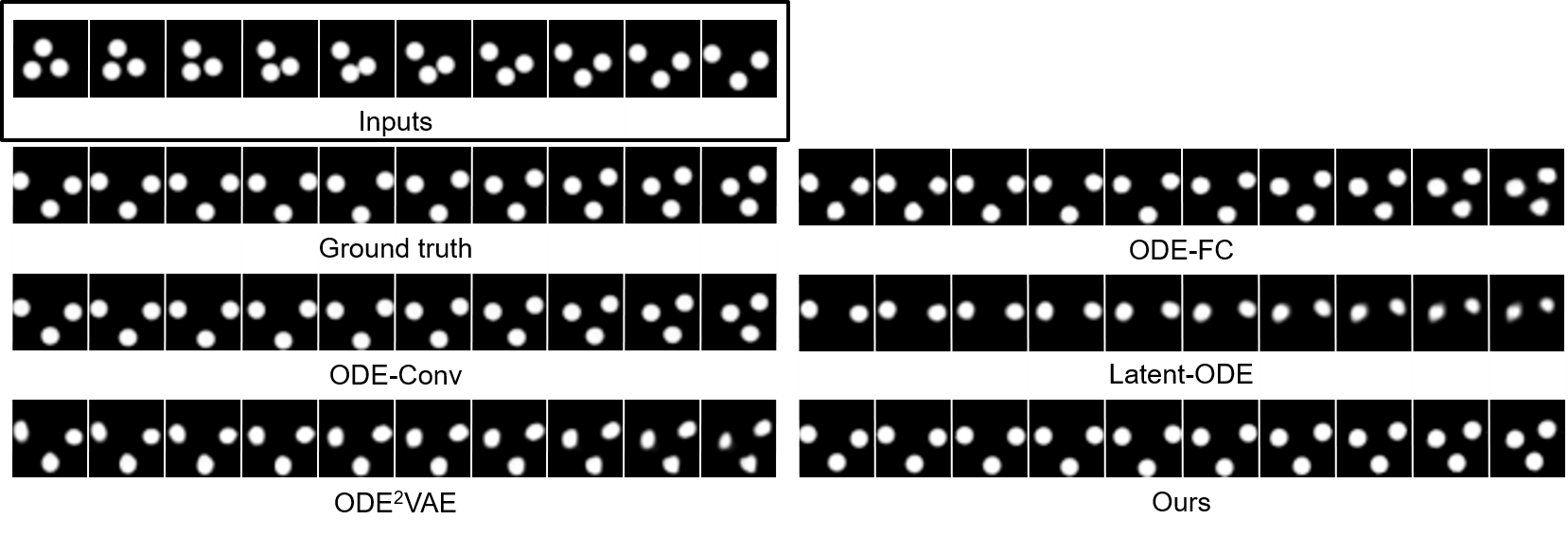}}
    \caption{Video extrapolation results of \model and baseline models from the Bouncing ball dataset.}
    \label{fig:base_extrap_bball}
\end{figure*}

\begin{table}[t!]
    \centering
    \footnotesize
    \begin{tabular}{clccc}
        \toprule
        \textbf{Datasets}
        &\multicolumn{1}{c}{\textbf{Model}}&
        SSIM$_{\uparrow}$ & LPIPS$_{\downarrow}$ & PSNR$_{\uparrow}$ \\
        \midrule
        \multirowcell{6}{\textit{KTH} \\ \textit{Action}}
        & ConvGRU
         & 0.764 &\ 0.379 &\ 21.52
        \\
        & PredNet
         & 0.825 &\ 0.242 &\ 22.15
        \\
        & DVF
         & 0.837 &\ 0.129 &\ 26.05
        \\
        & RCG
         & 0.820 &\ 0.187 &\ 21.92
        \\
        \cmidrule(lr){2-5}
        & \texttt{\model}
         & \textbf{0.878} &\ \textbf{0.080} &\ \textbf{28.19}
        \\
        \midrule
        \multirowcell{6}{\textit{Moving} \\ \textit{GIF}}
        & ConvGRU
         & 0.467 &\ 0.532 &\ 11.49
        \\
        & PredNet
         & 0.649 &\ 0.232 &\ 14.64
        \\
        & DVF
         & 0.777 &\ 0.215 &\ 16.39
        \\
        & RCG
         & 0.593 &\ 0.491 &\ 11.17
        \\
        \cmidrule(lr){2-5}
        & \texttt{\model}
         & \textbf{0.778} &\ \textbf{0.156} &\ \textbf{16.68}
        \\
        \midrule
        \multirowcell{6}{\textit{Penn} \\ \textit{Action}}
        & ConvGRU
         & 0.625 &\ 0.262 &\ 18.48
        \\
        & PredNet
         & 0.840 &\ 0.073 &\ 19.01
        \\
        & DVF
         & 0.790 &\ 0.102 &\ 21.90
        \\
        & RCG
         & 0.809 &\ 0.098 &\ 20.13
        \\
        \cmidrule(lr){2-5}
        & \texttt{\model}
         & \textbf{0.880} &\ \textbf{0.045} &\ \textbf{23.81}
        \\
        \bottomrule
    \end{tabular}
    \caption{Video extrapolation results}
    \label{Table:video extrapolation}
\end{table}

\noindent\textbf{Implementation Details.}
We employ Adamax~\cite{kingma2014adam}, a widely-used optimization method to iteratively train the ODE-based model.
We train \model for 500 epochs with a batch size of 8.
The learning rate is set initially as 0.001, then exponentially decaying at a rate of 0.99 per epoch.
In addition, we find that \model shows a slight performance improvement when the input frames are in reverse order.
A horizontal flip and a random rotation in the range of -10 to 10 degrees are used for data augmentation.
For the implementations of existing baselines, we follow the hyperparameters given in the original papers and conduct the experiments with the same number of epochs, the batch size, and data augmentation as our model.
For hyperparameters of \model, we use $\lambda_{\text{diff}}=1.0$, $\lambda_{\text{img}}=0.003$, and $\lambda_{\text{seq}}=0.003$. 
As for training ODEs, \model required only 7 hours for training on the KTH Action dataset using a single NVIDIA Titan RTX (using 6.5GB VRAM).

\begin{table}[b!]
    \centering
    \footnotesize
    \begin{tabular}{p{0.07\textwidth}<{\centering} p{0.07\textwidth}  p{0.05\textwidth}<{\centering} p{0.05\textwidth}<{\centering} p{0.05\textwidth}<{\centering}}
        \toprule
        \textbf{Datasets}&
        \multicolumn{1}{c}{\textbf{Model}}&
        SSIM$_{\uparrow}$ & LPIPS$_{\downarrow}$ & PSNR$_{\uparrow}$
        \\
        \midrule
        \multirowcell{4}{\textit{KTH} \\ \textit{Action}}
        & DVF
         & \textbf{0.954} &\ \textbf{0.037} &\ \textbf{36.28}
        \\
        & UVI
         & 0.934 &\ 0.055 &\ 29.97
        \\
        \cmidrule(lr){2-5}
        & \texttt{\model}
         & 0.911 &\ 0.048 &\ 31.77
        \\
        \midrule
        \multirowcell{4}{\textit{Moving} \\ \textit{GIF}}
        & DVF
         & \textbf{0.850} &\ 0.130 &\ \textbf{19.41}
        \\
        & UVI
         & 0.700 &\ 0.163 &\ 17.13
        \\
        \cmidrule(lr){2-5}
        & \texttt{\model}
         & 0.815 &\ \textbf{0.115} &\ 18.44
        \\
        \midrule
        \multirowcell{4}{\textit{Penn} \\ \textit{Action}}
        & DVF
         & \textbf{0.955} &\ \textbf{0.024} &\ \textbf{30.11}
        \\
        & UVI
         & 0.904 &\ 0.042 &\ 25.21
        \\
        \cmidrule(lr){2-5}
        & \texttt{\model}
         & 0.920 &\ 0.033 &\ 26.73
        \\
        \bottomrule
    \end{tabular}
    \caption{Video interpolation results. We compare \model(unsupervised) with DVF (supervised) and UVI (unupservised).}
    \label{Table:video interpolation}
\end{table}

\subsection{Comparison with Neural-ODE-based Models}
\noindent\textbf{Quantitative Comparison.} We compare \model against existing neural-ODE-based models such as ODE$^2$VAE~\cite{yildiz2019ode2vae} and latent ODE~\cite{rubanova2019latent}.
In addition, to verify the effectiveness of ODE-ConvGRU described in Section~\ref{sec:encoder}, we design the variants of neural ODEs (e.g., ODE-FC, ODE-Conv) by removing ODE-ConvGRU; thus, both take the channel-wise concatenated frames as an input for the Conv-Encoder $E$. We call the variants depending on the types of the derivative function $f_{\phi}$: fully-connected layers (ODE-FC) and convolutional layers (ODE-Conv).

Table~\ref{Table:2:comparison_ode} shows that \model significantly outperforms all other baselines both in interpolation and extrapolation tasks. This improvement can be attributed to the ODE-ConvGRU and the linear composition, which helps \model effectively maintain spatio-temporal information while preserving the sharpness of the outputs.
This is also supported by an observation that ODE-Conv outperforms ODE-FC, achieving higher scores by simply using convolutional layers to estimate the derivatives of hidden states where spatial information resides.
We find that the VAE architecture of ODE$^2$VAE and Latent ODE makes the training unstable, as the KL divergence loss of high dimensional representations does not converge well. Due to this, these models often fail to generate realistic images, resulting in suboptimal performance.
Qualitative results are reported in the supplementary material.

\begin{figure*}[t!]
    \centering
    \centerline{\includegraphics[width=\textwidth]{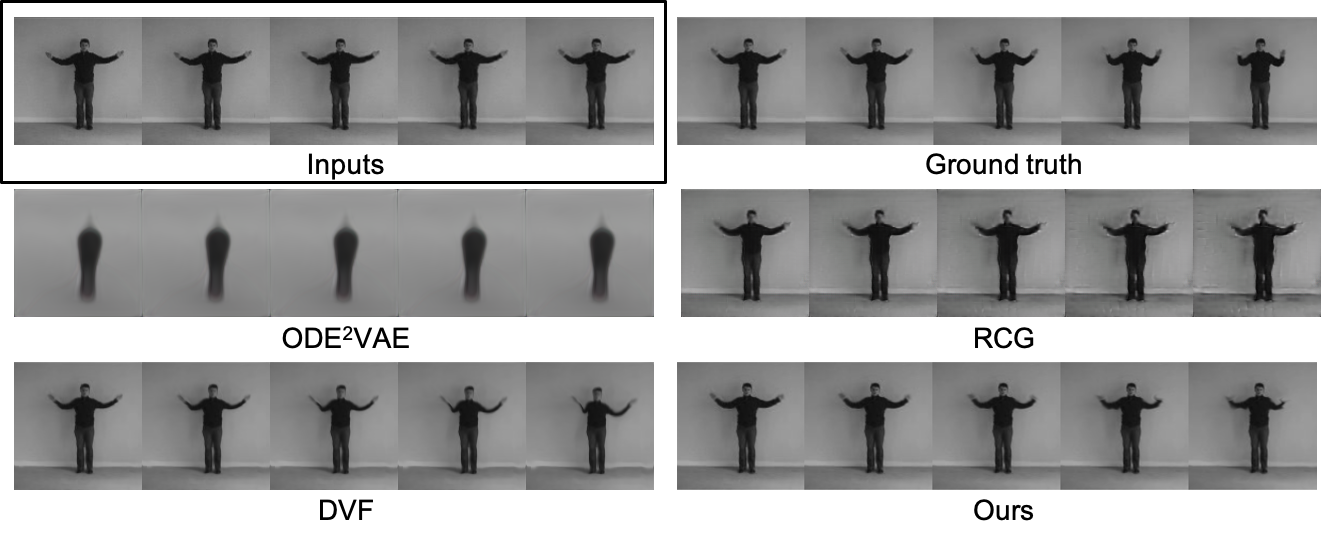}}
    \caption{Video extrapolation results of \model and baseline models from KTH-Action.}
    \label{fig:baseline model comparison}
\end{figure*}


\noindent\textbf{Non-linear Motion.} We validate the effectiveness of \model in handling a non-linear motion using bouncing ball dataset.
Given a sequence of 10 input frames, the model is trained to predict the next 10 future frames.
The motion of a ball often involves abrupt changes in its moving direction when it hits the wall, which can be viewed as a clear example of a non-linear motion.
As seen in the first half of the second row of Fig~\ref{fig:base_extrap_bball}, we can observe the non-linear, bouncing movement of a ball at the bottommost. Although such dynamics are non-linear, as shown in Fig~\ref{fig:base_extrap_bball}, our model successfully predicts the reflection.
Not surprisingly, most baselines work well, yet \model still outperforms the baselines, demonstrating its superiority even on a low-dimensional dataset.


\subsection{Comparison with Task-specific Models}
We compare the performance of \model against various state-of-the-art video interpolation~\cite{liu2017video, reda2019unsupervised} and extrapolation models~\cite{ballas2015delving, lotter2016deep, kwon2019predicting, liu2017video}.

\noindent\textbf{Video Extrapolation.}
As baselines, we adopt PredNet~\cite{lotter2016deep}, Deep Voxel Flow(DVF)~\cite{liu2017video}, Retrospective Cycle GAN(RCG)~\cite{kwon2019predicting}.
As shown in Table~\ref{Table:video extrapolation}, \model significantly outperforms all other baseline models in all metrics.
It is noteworthy that the performance gap is wider for Moving GIF, which contains more dynamic object movements (compared to rather slow movements in KTH-Action and Penn-Action), indicating \model's superior ability to learn complex dynamics.
Furthermore, qualitative comparison shown in Figure~\ref{fig:baseline model comparison} demonstrates that our model successfully learns the underlying dynamics of the object, and generates more realistic frames compared to baselines.
In summary, \model not only generates superior video frames compared to various state-of-the-art video extrapolation models, but also has the unique ability to generate frames at arbitrary timesteps.

\begin{figure}[t!]
    \centering
    \includegraphics[width=0.95\linewidth]{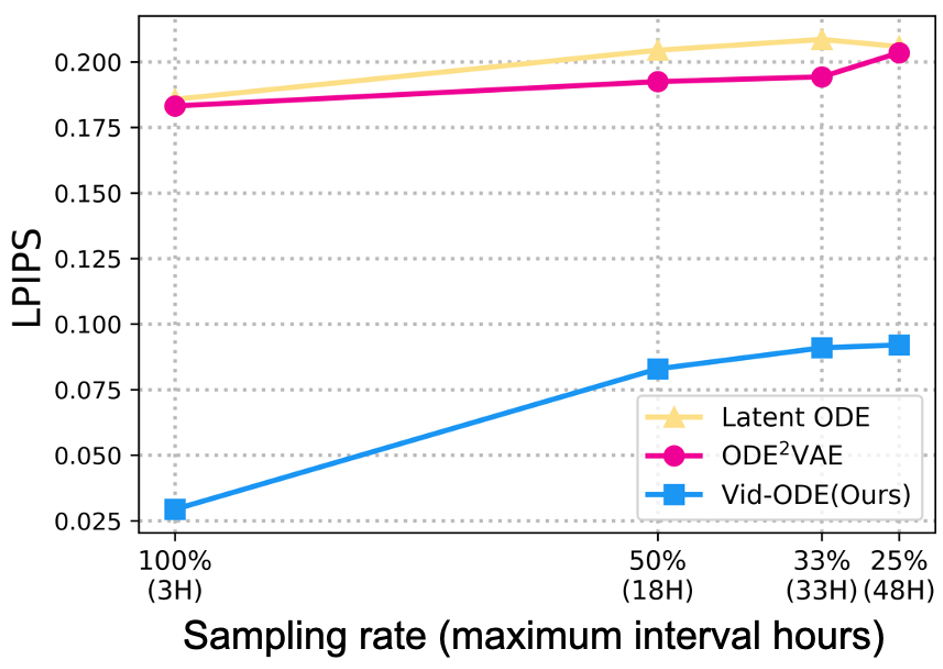}
    \caption{Change of LPIPS at different sampling rates. \model outperforms the baselines at all sampling rates and shows relatively small declines in performance as inputs are sparsely drawn.}
    \label{fig:MSE_graph}
\end{figure}

\begin{figure*}[t!]
    \begin{center}
    \centerline{\includegraphics[width=\linewidth]{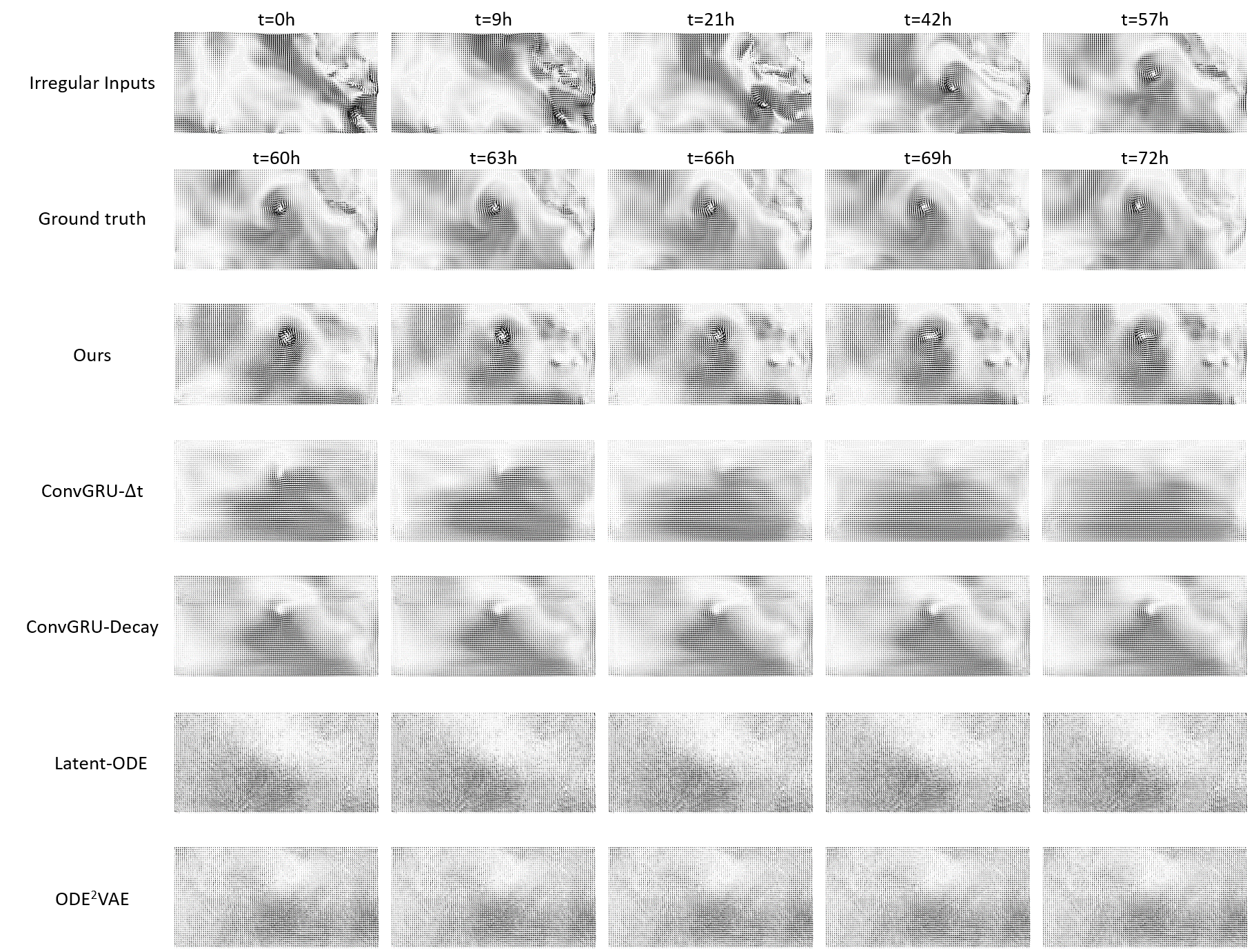}}
    \end{center}
    \caption{Qualitative comparisons with the baselines on the CAM5 hurricane dataset. For input, we sample 5 frames with different intervals (first row; \textit{i.e.,} 0, 9, 21, 42, and 57 hours). Given the irregularly sampled frames, every model including \model predicts the frames at regular timesteps (\textit{i.e.,} 60, 63, 66, 69, and 72 hours). For visualization, we use zonal wind (U850) and meridional wind (V850) together to represent the vector at each pixel. Our model outperforms the baselines, successfully synthesizing the shape and the trajectory of the hurricane.}
    \label{fig:base_extrap_hurricane}
\end{figure*}

\begin{figure}
    \centering
    \includegraphics[width=\linewidth]{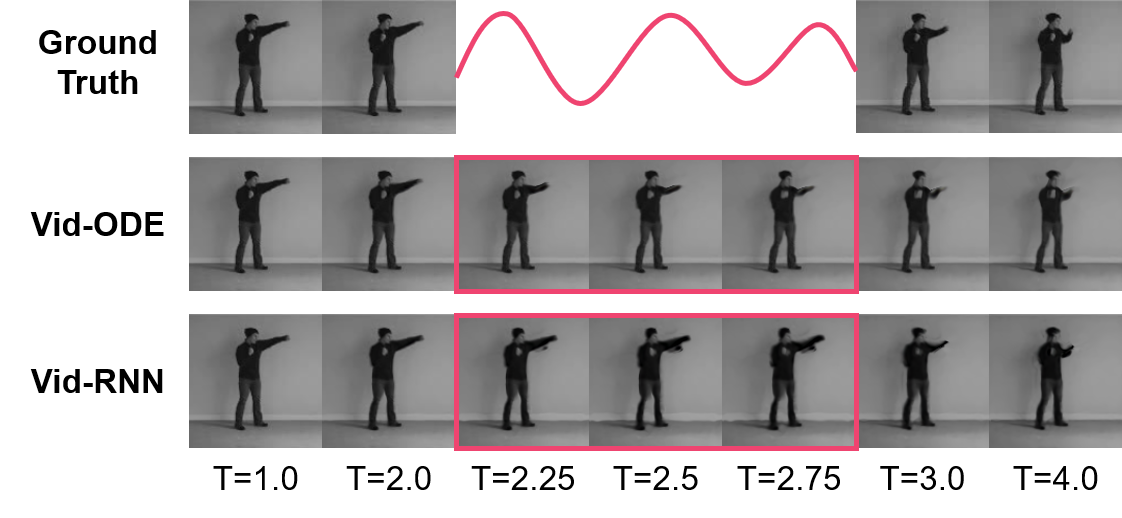}
    \caption{Comparison between \model and Vid-RNN, which has the same architecture as \model except all ODE components are replaced with RNNs, interpolating non-existing in-between frames at arbitrary timesteps.}
    \label{fig:baseline_continuous}
\end{figure}

\noindent\textbf{Video Interpolation.}
We compare \model with Unsupervised Video Interpolation (UVI)~\cite{reda2019unsupervised}, which is trained to interpolate in-between frames in an unsupervised manner.
We additionally compare with a supervised interpolation method, DVF~\cite{liu2017video}, to measure the headroom for potential further improvement.
As shown in Table~\ref{Table:video interpolation}, \model outperforms UVI in all cases (especially in Moving GIF), except for SSIM in KTH-Action. As expected, we see some gap between \model and the supervised approach (DVF).

\begin{table*}[t]
    \centering
    \footnotesize
    \centering
    \begin{tabular}{lcccccc}
        \toprule
        \multicolumn{1}{c}{\multirow{2}{*}{\textbf{Methods}}}
        & \multicolumn{3}{c}{\textbf{Video Interpolation}} & 
        \multicolumn{3}{c}{\textbf{Video Extrapolation}}\\
        \cmidrule(lr){2-4} \cmidrule(lr){5-7}
         & SSIM$_{\uparrow}$ & LPIPS$_{\downarrow}$ & PSNR$_{\uparrow}$ 
         & SSIM$_{\uparrow}$ & LPIPS$_{\downarrow}$ & PSNR$_{\uparrow}$  \\
        \midrule
        ($\mathtt{A}$) ODE-Conv
         & 0.769 &\ 0.416 &\ 25.12
         & 0.768 &\ 0.429 &\ 24.31
        \\
        ($\mathtt{B}$) Vanilla \texttt{\model}
         & 0.864 &\ 0.247 &\ 27.81
         & 0.853 &\ 0.262 &\ 26.49
        \\
        ($\mathtt{C}$) + Adversarial learning
         & 0.866 &\ 0.226 &\ 28.60
         & 0.856 &\ 0.245 &\ 27.69
        \\
        ($\mathtt{D}$) + Optical flow warping
         & \textbf{0.912} &\ 0.052 &\ 31.60
         & 0.862 &\ 0.085 &\ \textbf{28.30}
        \\
        ($\mathtt{E}$) + Mask composition
         & 0.911 &\ \textbf{0.048} &\ \textbf{31.77}
         & \textbf{0.878} &\ \textbf{0.080} &\ 28.19
         \\
        \bottomrule
    \end{tabular}
    \caption{Performance improvement by adding each component to the \model suggesting the applicability of each component to improve performance of tasks.}
    \label{Ablation Table}
\end{table*}

\begin{table}[h]
    \centering
    \footnotesize
    \centering
    \begin{tabular}{clcc}
        \toprule
        \textbf{Datasets}
        &\multicolumn{1}{c}{\textbf{Model}}&
        LPIPS$_{\downarrow}$ & MSE$_{\downarrow}$ \\
        \midrule
        \multirowcell{5}{\textit{CAM5}}
        & ConvGRU-$\Delta$t 
         & 0.270 &\ 2.439
        \\
        & ConvGRU-Decay
         & 0.160 &\ 0.583
        \\
        & Latent ODE
         & 0.206 &\ 0.554
        \\
        & ODE$^2$VAE
         & 0.203 &\ 0.551
        \\
        \cmidrule(lr){2-4}
        & \texttt{\model}
         & \textbf{0.092} &\ \textbf{0.515}
        \\
        \bottomrule
    \end{tabular}
    \caption{Extrapolation results for irregularly-sampled input video from the CAM5 hurricane dataset. MSE ($\times$ 10$^{-3}$)}
    \label{Table:irregular}
\end{table}

\noindent\textbf{Irregular Video Prediction.}
One of the distinguishing aspects of \model is its ability to handle videos of an arbitrary sampling rate.
We use CAM5 to test \model's ability to cope with irregularly sampled input, where we force the model to extrapolate at a higher rate (\textit{i.e.} every three hours) than the input's sampling rate.
We randomly sample 5 input frames from each hurricane video where the interval can be as large as 48 hours.
For baselines, we use Latent ODE, ODE$^2$VAE~\cite{yildiz2019ode2vae}, ConvGRU-$\Delta_t$, and ConvGRU-Decay where the last two were implemented by replacing the RNN cell of RNN-$\Delta_t$ and RNN-Decay~\cite{che2018recurrent} to ConvGRU~\cite{ballas2015delving}.
Table~\ref{Table:irregular} shows that \model outperforms baselines in both LPIPS and MSE, demonstrating the \model's ability to process irregularly sampled video frames.
Furthermore, visual comparison shown in Figure~\ref{fig:base_extrap_hurricane} demonstrates the capacity of \model to handle spatial-temporal dynamics from irregularly sampled inputs.
Lastly, we measure MSE and LPIPS on CAM5 dataset while changing the input’s sampling rate to evaluate the effect of irregularity. As shown in Figure~\ref{fig:MSE_graph}, all models perform worse as sampling rate decreases, demonstrating the difficulty of handling the sparsely sampled inputs. Still, \model outperforms the baselines at the irregularly sampled inputs as well as regularly sampled inputs ($i.e.,$ 100$\%$ sampling rate), which verify its superior capability to cope with irregular inputs.

\begin{figure}
    \begin{center}
    \centerline{\includegraphics[width=0.9\linewidth]{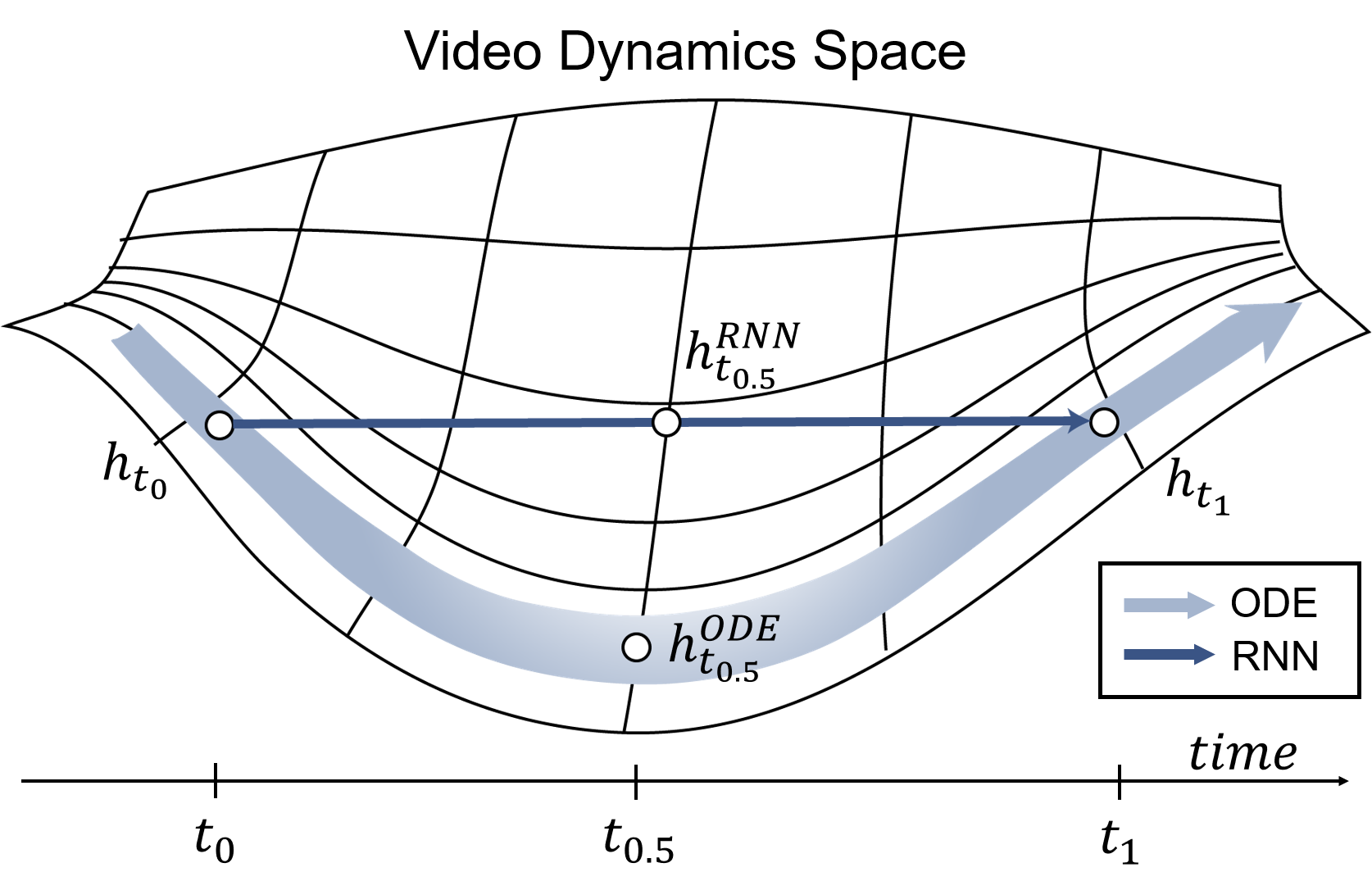}}
    \end{center}
    \caption{In the video dynamics space, RNN is limited to learn the representations only at observed timesteps ($t_{0}, t_{1}$). On the other hand, \model is encouraged to learn the entire geometry, enabling it to generate video frames at unseen timesteps ($t_{0.5}$).}
    \label{fig:space_figure}
\end{figure}

\subsection{Analysis of Individual Components}

\noindent\textbf{Need for Learning the Continuous Video Dynamics.}
To emphasize the need for learning the continuous video dynamics using the ODE, we compare \model to Vid-RNN, which replaces ODE components in both the encoder and decoder of \model with ConvGRU while retaining all other components such as linear composition.
Using Vid-RNN, we can obtain video representations at arbitrary timesteps by interpolating its decoder (ConvGRU) hidden states from two adjacent regular timesteps.
If Vid-RNN could generate video frames at unseen timesteps as well as \model, then an ODE-based video generation would be unnecessary.
However, Figure~\ref{fig:baseline_continuous} shows that is not the case. 
While \model is successfully inferring video frames at unseen timesteps ($t=2.25, t=2.5, t=2.75$) thanks to learning the underlying video dynamics, Vid-RNN generates unrealistic video frames due to simply blending two adjacent latent representations.
The intuition behind such behavior is described in Figure~\ref{fig:space_figure}, where the advantage of the ODE-based approach is evident when handling continuous time.

\noindent\textbf{Ablation Study.}
Table~\ref{Ablation Table} depicts the effectiveness of each component of the \model.
Starting with a simple baseline ODE-Conv $\mathtt{(A)}$, we first compare vanilla \model $\mathtt{(B)}$ equipped with the proposed ODE-ConvGRU.
We see a significant boost in performance, meaning that the ODE-ConvGRU cell better captures spatio-temporal dynamics in the video and is able to generate high-quality frames.
As depicted by $\mathtt{(C)}$ in Table~\ref{Ablation Table}, adding the adversarial loss to $\mathtt{(B)}$ improves performance, especially in LPIPS, suggesting that the image and sequence discriminators help the model generate realistic images.
From $\mathtt{(C)}$ to $\mathtt{(D)}$, we add the optical flow warping (Eq.~\eqref{eq:decoder}), which significantly enhances the performance by effectively learning the video dynamics.
As the last component, we add the linear composition $\Psi$ ($\mathtt{E}$).
The performance boost from $\mathtt{(D)}$ to $\mathtt{(E)}$ might seem marginal.
Comparing the warped image with the final product in Figure~\ref{fig:mask_visualization}, however, demonstrates that using the image difference to \textit{fill in} the disappeared pixels in the warped image indeed enhances the visual quality of the output.

\section{Conclusions}
In this paper, we propose \model which enjoys the continuous nature of neural ODEs to generate video frames at any given timesteps.
Combining the ODE-ConvGRU with the linear composition of optical flow and image difference, \model successfully demonstrates its ability to generate high-quality video frames in the continuous-time domain using four real-world video datasets for both video interpolation and video extrapolation.
Despite its success in continuous-time video generation, \model tends to yield degraded outcomes as the number of predictions increases because of its autoregressive nature.
In future work, we plan to study how to adopt a flexible structure to address this issue.

\section*{Acknowledgements}
This work was partly supported by Institute of Information \& communications Technology Planning \& Evaluation (IITP) grant funded by the Korea government(MSIT) (No. 2019-0-00075, Artificial Intelligence Graduate School Program(KAIST) and No. 2020-0-00368, A Neural-Symbolic Model for Knowledge Acquisition and Inference Techniques) and by the National Research Foundation of Korea (NRF) grant funded by the Korean government (MSIT) (No. NRF-2018M3E3A1057305).

\section*{Ethical Impact}
Our proposed \model model learns the continuous flow of videos from a sequence of frames from potentially irregular training data and is capable of synthesizing new frames at any given timesteps. Our framework is applicable to the broad scope of spatio-temporal data without limiting to multi-media data. For example, as discussed in the main sections, this can be especially useful for scientific data where the assumption of regularly sampled time step does not always hold. Specifically, the application of the proposed model to climate data, where the measurement is costly and sparse while it is sometimes beneficial to forecast at the denser rate, is critical. This can potentially bring a significant impact on weather forecasting with improved estimation quality as well as with less cost compared to traditional scientific models. For instance, in disaster prevention plans for extreme climate events, the decision-makers often rely on simulation or observation data with sparse timesteps, which is only available out there. This limits the capability to forecast in more frequent timesteps and thus prevent solid decisions based on accurate disaster scenarios. Our \model takes a significant step towards a fully data-driven approach to forecasting extreme climate events by addressing this issue.

\small
\bibstyle{unsrt}
\bibliography{reference}

\begin{thebibliography}{34}
\providecommand{\natexlab}[1]{#1}
\providecommand{\url}[1]{\texttt{#1}}
\providecommand{\urlprefix}{URL }
\expandafter\ifx\csname urlstyle\endcsname\relax
  \providecommand{\doi}[1]{doi:\discretionary{}{}{}#1}\else
  \providecommand{\doi}{doi:\discretionary{}{}{}\begingroup
  \urlstyle{rm}\Url}\fi

\bibitem[{Ballas et~al.(2015)Ballas, Yao, Pal, and
  Courville}]{ballas2015delving}
Ballas, N.; Yao, L.; Pal, C.; and Courville, A. 2015.
\newblock Delving deeper into convolutional networks for learning video
  representations.
\newblock \emph{arXiv:1511.06432} .

\bibitem[{Bao et~al.(2019)Bao, Lai, Ma, Zhang, Gao, and Yang}]{bao2019depth}
Bao, W.; Lai, W.-S.; Ma, C.; Zhang, X.; Gao, Z.; and Yang, M.-H. 2019.
\newblock Depth-aware video frame interpolation.
\newblock In \emph{Proc. of the IEEE Conference on Computer Vision and Pattern
  Recognition (CVPR)}.

\bibitem[{Che et~al.(2018)Che, Purushotham, Cho, Sontag, and
  Liu}]{che2018recurrent}
Che, Z.; Purushotham, S.; Cho, K.; Sontag, D.; and Liu, Y. 2018.
\newblock Recurrent neural networks for multivariate time series with missing
  values.
\newblock \emph{Scientific reports} 8(1): 1--12.

\bibitem[{Chen et~al.(2018)Chen, Rubanova, Bettencourt, and
  Duvenaud}]{chen2018neural}
Chen, T.~Q.; Rubanova, Y.; Bettencourt, J.; and Duvenaud, D.~K. 2018.
\newblock Neural ordinary differential equations.
\newblock In \emph{Advances in Neural Information Processing Systems (NIPS)}.

\bibitem[{De~Brouwer et~al.(2019)De~Brouwer, Simm, Arany, and
  Moreau}]{de2019gru}
De~Brouwer, E.; Simm, J.; Arany, A.; and Moreau, Y. 2019.
\newblock {GRU}-{ODE}-{B}ayes: Continuous modeling of sporadically-observed
  time series.
\newblock In \emph{Advances in Neural Information Processing Systems (NIPS)}.

\bibitem[{Denton and Fergus(2018)}]{denton2018stochastic}
Denton, E.; and Fergus, R. 2018.
\newblock Stochastic Video Generation with a Learned Prior.
\newblock In \emph{International Conference on Machine Learning}, 1174--1183.

\bibitem[{Dosovitskiy et~al.(2015)Dosovitskiy, Fischer, Ilg, Hausser, Hazirbas,
  Golkov, Van Der~Smagt, Cremers, and Brox}]{dosovitskiy2015flownet}
Dosovitskiy, A.; Fischer, P.; Ilg, E.; Hausser, P.; Hazirbas, C.; Golkov, V.;
  Van Der~Smagt, P.; Cremers, D.; and Brox, T. 2015.
\newblock {F}low{N}et: Learning optical flow with convolutional networks.
\newblock In \emph{Proc. of the IEEE International Conference on Computer
  Vision (ICCV)}.

\bibitem[{Dupont, Doucet, and Teh(2019)}]{dupont2019augmented}
Dupont, E.; Doucet, A.; and Teh, Y.~W. 2019.
\newblock Augmented neural {ODE}s.
\newblock In \emph{Advances in Neural Information Processing Systems (NIPS)}.

\bibitem[{Franceschi et~al.(2020)Franceschi, Delasalles, Chen, Lamprier, and
  Gallinari}]{franceschi2020stochastic}
Franceschi, J.-Y.; Delasalles, E.; Chen, M.; Lamprier, S.; and Gallinari, P.
  2020.
\newblock Stochastic Latent Residual Video Prediction.
\newblock \emph{arXiv preprint arXiv:2002.09219} .

\bibitem[{Gao et~al.(2019)Gao, Xu, Cai, Wang, Yu, and
  Darrell}]{gao2019disentangling}
Gao, H.; Xu, H.; Cai, Q.-Z.; Wang, R.; Yu, F.; and Darrell, T. 2019.
\newblock Disentangling propagation and generation for video prediction.
\newblock In \emph{Proc. of the IEEE International Conference on Computer
  Vision (ICCV)}.

\bibitem[{Hao, Huang, and Belongie(2018)}]{hao2018controllable}
Hao, Z.; Huang, X.; and Belongie, S. 2018.
\newblock Controllable video generation with sparse trajectories.
\newblock In \emph{Proc. of the IEEE Conference on Computer Vision and Pattern
  Recognition (CVPR)}.

\bibitem[{Ilg et~al.(2017)Ilg, Mayer, Saikia, Keuper, Dosovitskiy, and
  Brox}]{ilg2017flownet}
Ilg, E.; Mayer, N.; Saikia, T.; Keuper, M.; Dosovitskiy, A.; and Brox, T. 2017.
\newblock Flownet 2.0: Evolution of optical flow estimation with deep networks.
\newblock In \emph{Proc. of the IEEE Conference on Computer Vision and Pattern
  Recognition (CVPR)}.

\bibitem[{Jiang et~al.(2018)Jiang, Sun, Jampani, Yang, Learned-Miller, and
  Kautz}]{jiang2018super}
Jiang, H.; Sun, D.; Jampani, V.; Yang, M.-H.; Learned-Miller, E.; and Kautz, J.
  2018.
\newblock Super slomo: High quality estimation of multiple intermediate frames
  for video interpolation.
\newblock In \emph{Proc. of the IEEE Conference on Computer Vision and Pattern
  Recognition (CVPR)}.

\bibitem[{Kim et~al.(2019)Kim, Park, Chung, Lee, Lee, Kim, Prabhat, and
  Choo}]{kim2019}
Kim, S.; Park, S.; Chung, S.; Lee, J.; Lee, Y.; Kim, H.; Prabhat, M.; and Choo,
  J. 2019.
\newblock Learning to Focus and Track Extreme Climate Events.
\newblock In \emph{Proc. of the British Machine Vision Conference (BMVC)}.

\bibitem[{Kingma and Ba(2014)}]{kingma2014adam}
Kingma, D.~P.; and Ba, J. 2014.
\newblock Adam: A method for stochastic optimization.
\newblock \emph{arXiv:1412.6980} .

\bibitem[{Kwon and Park(2019)}]{kwon2019predicting}
Kwon, Y.-H.; and Park, M.-G. 2019.
\newblock Predicting future frames using retrospective cycle {GAN}.
\newblock In \emph{Proc. of the IEEE Conference on Computer Vision and Pattern
  Recognition (CVPR)}.

\bibitem[{Lee et~al.(2018)Lee, Zhang, Ebert, Abbeel, Finn, and
  Levine}]{lee2018stochastic}
Lee, A.~X.; Zhang, R.; Ebert, F.; Abbeel, P.; Finn, C.; and Levine, S. 2018.
\newblock Stochastic adversarial video prediction.
\newblock \emph{arXiv preprint arXiv:1804.01523} .

\bibitem[{Liang et~al.(2017)Liang, Lee, Dai, and Xing}]{liang2017dual}
Liang, X.; Lee, L.; Dai, W.; and Xing, E.~P. 2017.
\newblock Dual motion GAN for future-flow embedded video prediction.
\newblock In \emph{Proc. of the IEEE International Conference on Computer
  Vision (ICCV)}.

\bibitem[{Liu et~al.(2017)Liu, Yeh, Tang, Liu, and Agarwala}]{liu2017video}
Liu, Z.; Yeh, R.~A.; Tang, X.; Liu, Y.; and Agarwala, A. 2017.
\newblock Video frame synthesis using deep voxel flow.
\newblock In \emph{Proc. of the IEEE International Conference on Computer
  Vision (ICCV)}.

\bibitem[{Lotter, Kreiman, and Cox(2016)}]{lotter2016deep}
Lotter, W.; Kreiman, G.; and Cox, D. 2016.
\newblock Deep predictive coding networks for video prediction and unsupervised
  learning.
\newblock \emph{arXiv:1605.08104} .

\bibitem[{Mao et~al.(2017)Mao, Li, Xie, Lau, Wang, and
  Paul~Smolley}]{mao2017least}
Mao, X.; Li, Q.; Xie, H.; Lau, R.~Y.; Wang, Z.; and Paul~Smolley, S. 2017.
\newblock Least squares generative adversarial networks.
\newblock In \emph{Proc. of the IEEE International Conference on Computer
  Vision (ICCV)}.

\bibitem[{Reda et~al.(2019)Reda, Sun, Dundar, Shoeybi, Liu, Shih, Tao, Kautz,
  and Catanzaro}]{reda2019unsupervised}
Reda, F.~A.; Sun, D.; Dundar, A.; Shoeybi, M.; Liu, G.; Shih, K.~J.; Tao, A.;
  Kautz, J.; and Catanzaro, B. 2019.
\newblock Unsupervised Video Interpolation Using Cycle Consistency.
\newblock In \emph{Proc. of the IEEE International Conference on Computer
  Vision (ICCV)}.

\bibitem[{Revaud et~al.(2015)Revaud, Weinzaepfel, Harchaoui, and
  Schmid}]{revaud2015epicflow}
Revaud, J.; Weinzaepfel, P.; Harchaoui, Z.; and Schmid, C. 2015.
\newblock Epicflow: Edge-preserving interpolation of correspondences for
  optical flow.
\newblock In \emph{Proc. of the IEEE Conference on Computer Vision and Pattern
  Recognition (CVPR)}.

\bibitem[{Rubanova, Chen, and Duvenaud(2019)}]{rubanova2019latent}
Rubanova, Y.; Chen, T.~Q.; and Duvenaud, D.~K. 2019.
\newblock Latent Ordinary Differential Equations for Irregularly-Sampled Time
  Series.
\newblock In \emph{Advances in Neural Information Processing Systems (NIPS)}.

\bibitem[{Schuldt, Laptev, and Caputo(2004)}]{schuldt2004}
Schuldt, C.; Laptev, I.; and Caputo, B. 2004.
\newblock Recognizing human actions: a local SVM approach.
\newblock In \emph{Proc. of the International Conference on Pattern Recognition
  (ICPR)}.

\bibitem[{Siarohin et~al.(2019)Siarohin, Lathuili{\`e}re, Tulyakov, Ricci, and
  Sebe}]{siarohin2019animating}
Siarohin, A.; Lathuili{\`e}re, S.; Tulyakov, S.; Ricci, E.; and Sebe, N. 2019.
\newblock Animating arbitrary objects via deep motion transfer.
\newblock In \emph{Proc. of the IEEE Conference on Computer Vision and Pattern
  Recognition (CVPR)}.

\bibitem[{Wang et~al.(2019{\natexlab{a}})Wang, Jiang, Yang, Li, Long, and
  Fei-Fei}]{wang2018eidetic}
Wang, Y.; Jiang, L.; Yang, M.-H.; Li, L.-J.; Long, M.; and Fei-Fei, L.
  2019{\natexlab{a}}.
\newblock Eidetic 3d {LSTM}: A model for video prediction and beyond.
\newblock In \emph{Proc. of the International Conference on Learning
  Representations (ICLR)}.

\bibitem[{Wang et~al.(2017)Wang, Long, Wang, Gao, and Philip}]{wang2017predrnn}
Wang, Y.; Long, M.; Wang, J.; Gao, Z.; and Philip, S.~Y. 2017.
\newblock Predrnn: Recurrent neural networks for predictive learning using
  spatiotemporal lstms.
\newblock In \emph{Advances in Neural Information Processing Systems (NIPS)}.

\bibitem[{Wang et~al.(2019{\natexlab{b}})Wang, Zhang, Zhu, Long, Wang, and
  Yu}]{wang2019memory}
Wang, Y.; Zhang, J.; Zhu, H.; Long, M.; Wang, J.; and Yu, P.~S.
  2019{\natexlab{b}}.
\newblock Memory In Memory: A Predictive Neural Network for Learning
  Higher-Order Non-Stationarity from Spatiotemporal Dynamics.
\newblock In \emph{Proc. of the IEEE Conference on Computer Vision and Pattern
  Recognition (CVPR)}.

\bibitem[{Wang et~al.(2004)Wang, Bovik, Sheikh, and Simoncelli}]{wang2004image}
Wang, Z.; Bovik, A.~C.; Sheikh, H.~R.; and Simoncelli, E.~P. 2004.
\newblock Image quality assessment: from error visibility to structural
  similarity.
\newblock \emph{IEEE transactions on image processing} 13(4): 600--612.

\bibitem[{Xingjian et~al.(2015)Xingjian, Chen, Wang, Yeung, Wong, and
  Woo}]{xingjian2015convolutional}
Xingjian, S.; Chen, Z.; Wang, H.; Yeung, D.-Y.; Wong, W.-K.; and Woo, W.-c.
  2015.
\newblock Convolutional {LSTM} network: A machine learning approach for
  precipitation nowcasting.
\newblock In \emph{Advances in Neural Information Processing Systems (NIPS)}.

\bibitem[{Yildiz, Heinonen, and Lahdesmaki(2019)}]{yildiz2019ode2vae}
Yildiz, C.; Heinonen, M.; and Lahdesmaki, H. 2019.
\newblock {ODE}$^2${VAE}: Deep generative second order ODEs with Bayesian
  neural networks.
\newblock In \emph{Advances in Neural Information Processing Systems (NIPS)}.

\bibitem[{Zhang et~al.(2018)Zhang, Isola, Efros, Shechtman, and
  Wang}]{zhang2018unreasonable}
Zhang, R.; Isola, P.; Efros, A.~A.; Shechtman, E.; and Wang, O. 2018.
\newblock The unreasonable effectiveness of deep features as a perceptual
  metric.
\newblock In \emph{Proc. of the IEEE Conference on Computer Vision and Pattern
  Recognition (CVPR)}.

\bibitem[{Zhang, Zhu, and Derpanis(2013)}]{zhang2013}
Zhang, W.; Zhu, M.; and Derpanis, K.~G. 2013.
\newblock From actemes to action: {A} strongly-supervised representation for
  detailed action understanding.
\newblock In \emph{Proc. of the {IEEE} International Conference on Computer
  Vision (ICCV)}.

\end{thebibliography}

\newpage

\twocolumn[
\begin{center}
    \vspace*{2cm}
    \huge{\bf{Supplementary Material}}
    \vspace*{3cm}
\end{center}]

\section*{A. Implementation Details}
\label{sec:supp_impdetails}
We employ Adamax~\cite{kingma2014adam}, a widely-used optimization method to iteratively train the ODE-based model.
We train \model for 500 epochs with a batch size of 8.
The learning rate is set initially as 0.001, then exponentially decaying at a rate of 0.99 per epoch.
In addition, we find that \model shows a slight performance improvement when the input frames are in a reverse order.
A horizontal flip and a random rotation in the range of -10 to 10 degrees are used for data augmentation.
For the implementations of existing baselines, we follow the hyperparameters given in the original papers and conduct the experiments with the same number of epochs, the batch size and data augmentation as our model.
For hyperparameters of \model, we use $\lambda_{\text{diff}}=1.0$, $\lambda_{\text{img}}=0.003$, and $\lambda_{\text{seq}}=0.003$. 
As for training ODEs, \model required only 7 hours for training on KTH Action dataset using a single NVIDIA Titan RTX (using 6.5GB VRAM).

\subsection{Adversarial Loss for Interpolation}
\label{sec:supp_advloss}
For video interpolation, we make the generated sequence $\hat{\mathcal{X}}_\mathcal{S}$ for all timesteps by alternatively substituting a real frame in $\mathcal{X}_\mathcal{T}$ with a fake frame $\hat{\mathbf{X}}_{s_i}$. Formally, $\mathcal{L}^{\text{img}}_{\text{adv}}$ and $\mathcal{L}^{\text{seq}}_{\text{adv}}$ are formulated as
{ \small
\begin{align}
    &\min_{\footnotesize \text{\model}} \max_{D_{\text{img}}}
    \mathcal{L}^{\text{img}}_{\text{adv}} = 
    \mathbb{E}_{\mathbf{X}_{s_i} \sim p(\mathcal{X}_\mathcal{S})}
    \left[ \left( D_{\text{img}} \left(\mathbf{X}_{s_i} \right) - 1 \right)^2 
    \right] \nonumber \\
    & \hspace{2.2cm} + \mathbb{E}_{\mathcal{X}_\mathcal{T} \sim p(\mathcal{X}_\mathcal{T})}
    \left[ \left( D_{\text{img}}({\footnotesize \text{\model}}(\hat{\mathbf{X}}_{s_i}|\mathcal{X}_\mathcal{T})) \right)^2 \right] \nonumber \\
    &\min_{\footnotesize \text{\model}} \max_{D_{\text{seq}}}
    \mathcal{L}^{\text{seq}}_{\text{adv}} = 
    \mathbb{E}_{\mathcal{X}_\mathcal{T} \sim p(\mathcal{X}_\mathcal{T})}
    \left[ \left( D_{\text{seq}}(\mathcal{X}_\mathcal{T})-1 \right)^2 \right]
    \nonumber \\
    &+ \mathbb{E}_{\mathcal{X}_\mathcal{T} \sim p(\mathcal{X}_\mathcal{T})}
    \left[ \left(D_{\text{seq}} \big(\mathcal{X}_{t_0:t_{i-1}};{\footnotesize \text{\model}}(\hat{\mathbf{X}}_{s_i}|\mathcal{X}_\mathcal{T});\mathcal{X}_{t_{i+1}:t_{L}}\big)\right)^2 \right],
    \label{eq:adv_loss}
\end{align} }
where $\mathcal{X}_{t_0 :t_{i-1}};{\footnotesize \text{\model}}(\hat{\mathbf{X}}_{s_i}|\mathcal{X}_\mathcal{T});\mathcal{X}_{t_{i+1}:t_{L}}$ denotes a modified input sequence in which an intermediate frame $\mathbf{X}_{t_i}$ is replaced by the predicted frame $\hat{\mathbf{X}}_{s_i}$ (Recall that $s_i = t_{i}$ for interpolation).
Note that $\mathcal{X}_{t_{0}:t_{0}}=\mathcal{X}_{t_{L}:t_{L}}=\emptyset$, meaning that concatenation only occurs at backward or forward when $i=1$ and $i=L-1$, respectively.

\subsection{Model Architecture}
\label{sec:supp_arch}

\model architectures are shown in Tables~\ref{Encoder Arch}-\ref{Discriminator Arch}. The notations used are as follows; N: the number of the output channels, K: the kernel size, S: the stride size, P: the padding size, BN: batch normalization, Up: bilinear upsampling $\times 2$.

\section*{B. Architectures of Vid-ODE and Vid-RNN}
\label{sec:supp_rnncomp}

\begin{figure}[h!]
    \centering
    \includegraphics[width=\linewidth]{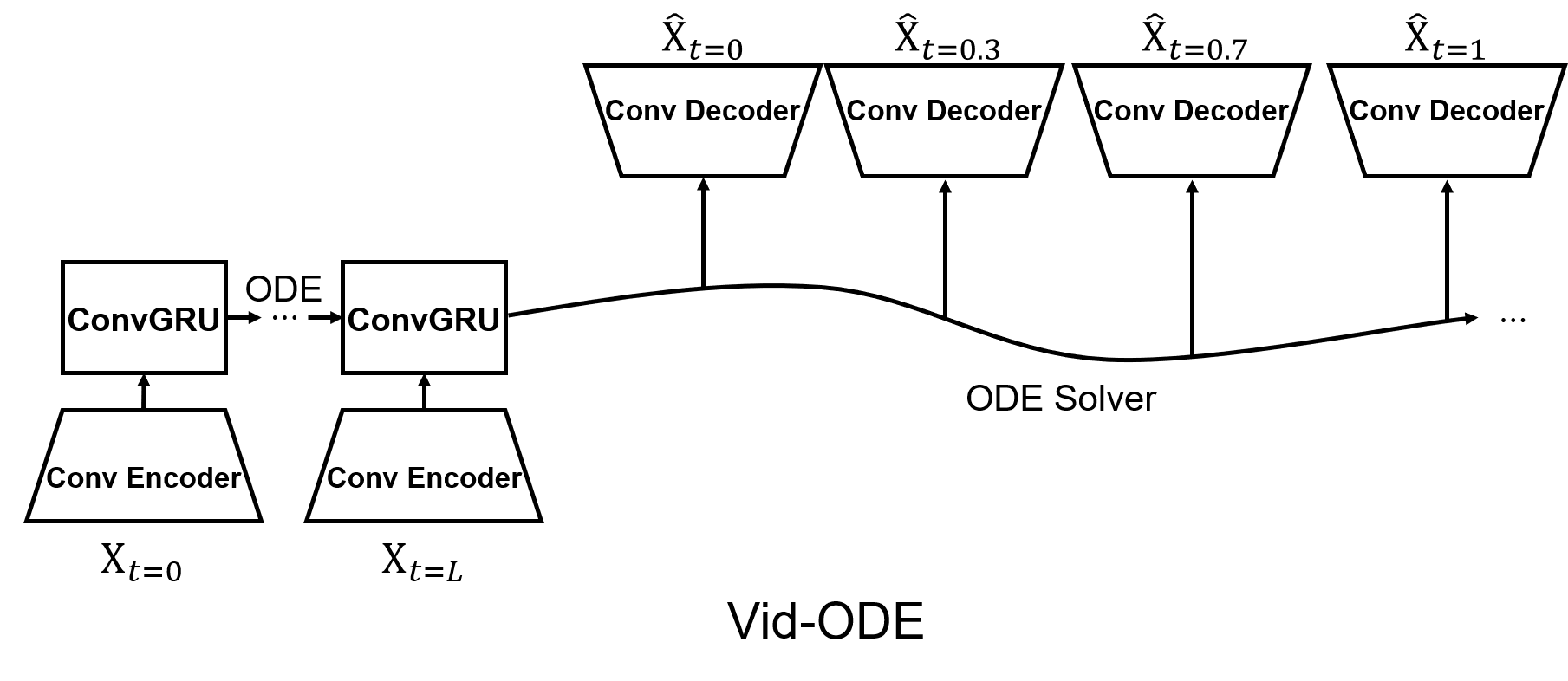}
    \includegraphics[width=\linewidth]{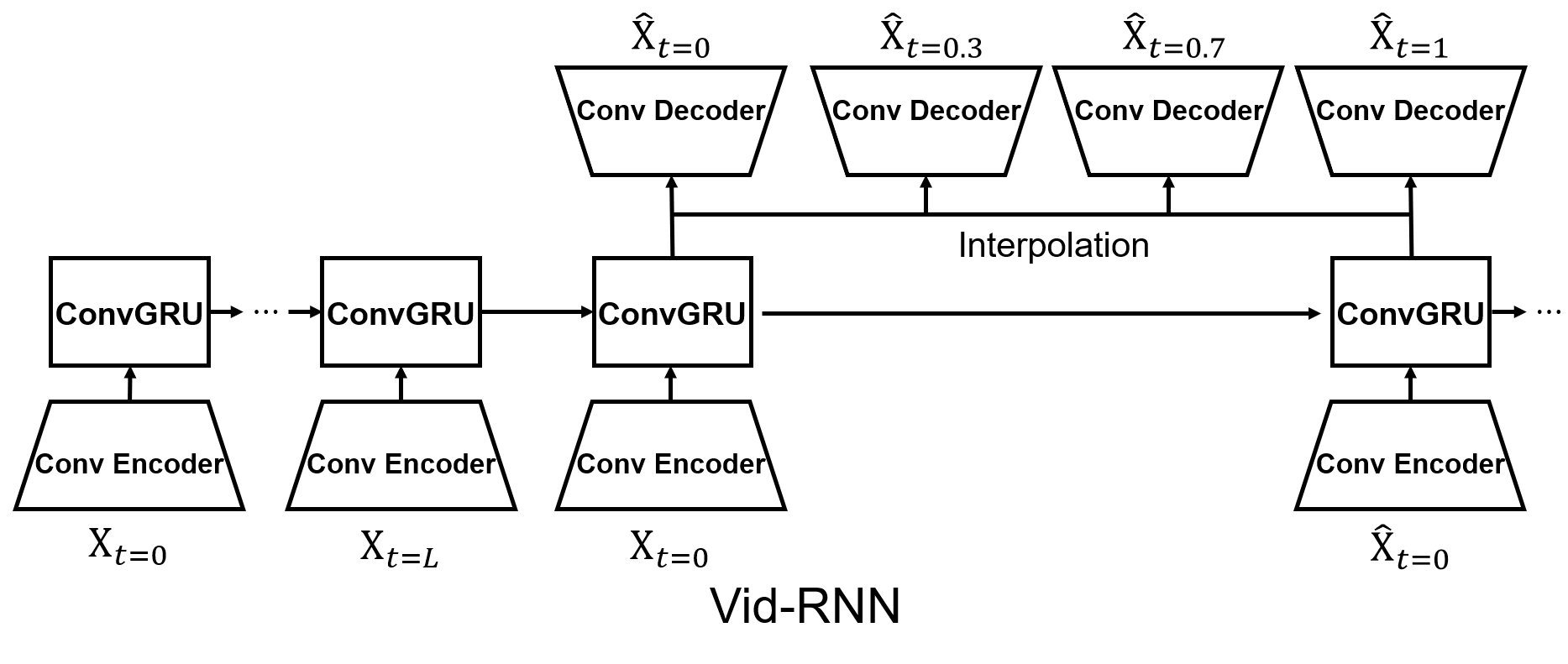}
    \caption{The architectures of \model and Vid-RNN for continuous video interpolation. 
    The main difference between \model and Vid-RNN lies in how to obtain the representations at unseen timesteps.
    In order to generate the frames at unseen timesteps, we interpolate the hidden states of Vid-RNN decoder from two observed timesteps. 
    The unseen frames are not used during training of Vid-RNN to make the same settings as \model.}
    \label{fig:model_vid_rnn}
\end{figure}

We compare \model and Vid-RNN to demonstrate the superior capability of the ODE component to learn continuous video dynamics. 
Vid-RNN is built by replacing ODE components in \model with ConvGRU while retaining all other components such as adversarial losses and the linear composition.
Figure~\ref{fig:model_vid_rnn} shows the the architectures of \model and Vid-RNN.
To clearly illustrate the main components of the architectures, we omit the detailed components of decoding phase such as optical flow and linear composition.

\section*{C. Additional Experiments}
\label{sec:supp_qualcomp}

This section presents additional experiments as follows:
\begin{itemize}
    \item Quantitative comparison results against the stochastic video prediction baselines.
    \item Quantitative comparison results given irregularly sampled inputs at different sampling rates.
    \item Qualitative comparison results with the video interpolation and extrapolation baselines.
    \item Continuous video generation results in varying time intervals based on 5-FPS input videos.
    \item High-resolution (\textit{e.g.,} $256 \times 256$) video interpolation and extrapolation results on Penn Action dataset.
\end{itemize}

\noindent\textbf{Additional Comparison with Baselines.}
In order to further evaluate our method, we additionally compare \model with SVG-FP, SVG-LP~\cite{denton2018stochastic}, SAVP~\cite{lee2018stochastic}, and SRVP~\cite{franceschi2020stochastic} for extrapolation. 
We used the official codes provided by the authors and followed the hyperparameters as presented in the official codes. 
Table~\ref{tab:comparison} shows \model consistently outperforms these existing approaches (except for one metric).

\begin{figure*}[h]
    \centering
    \centerline{\includegraphics[width=0.7\linewidth]{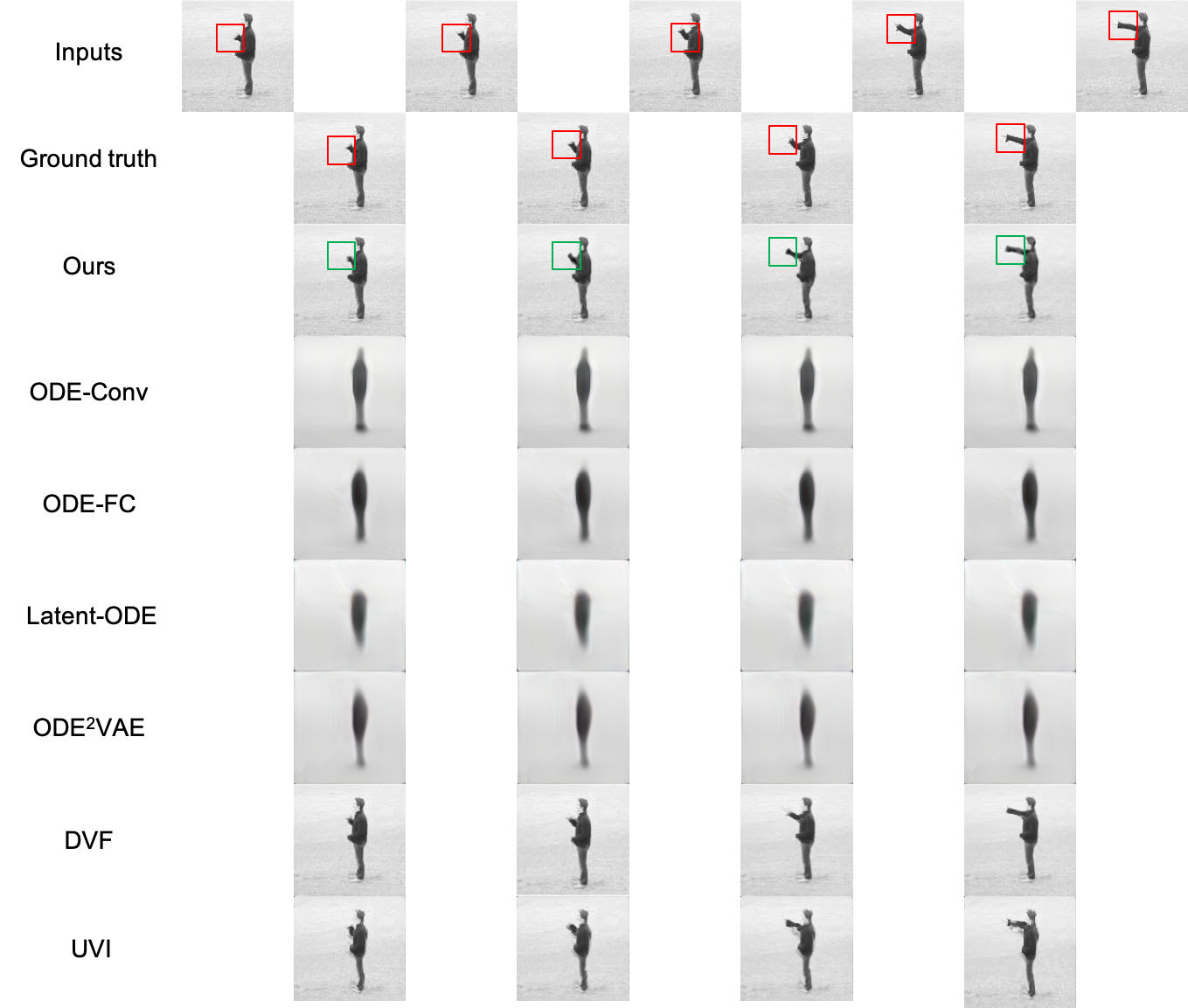}}
    \vspace{-0.1cm}
    \caption{Qualitative comparisons with interpolation baselines on the KTH Action dataset.}
    \label{fig:base_interp_kth}
    \vspace{-0.5cm}
\end{figure*}

\begin{figure*}[h]
    \centering
    \centerline{\includegraphics[width=0.7\linewidth]{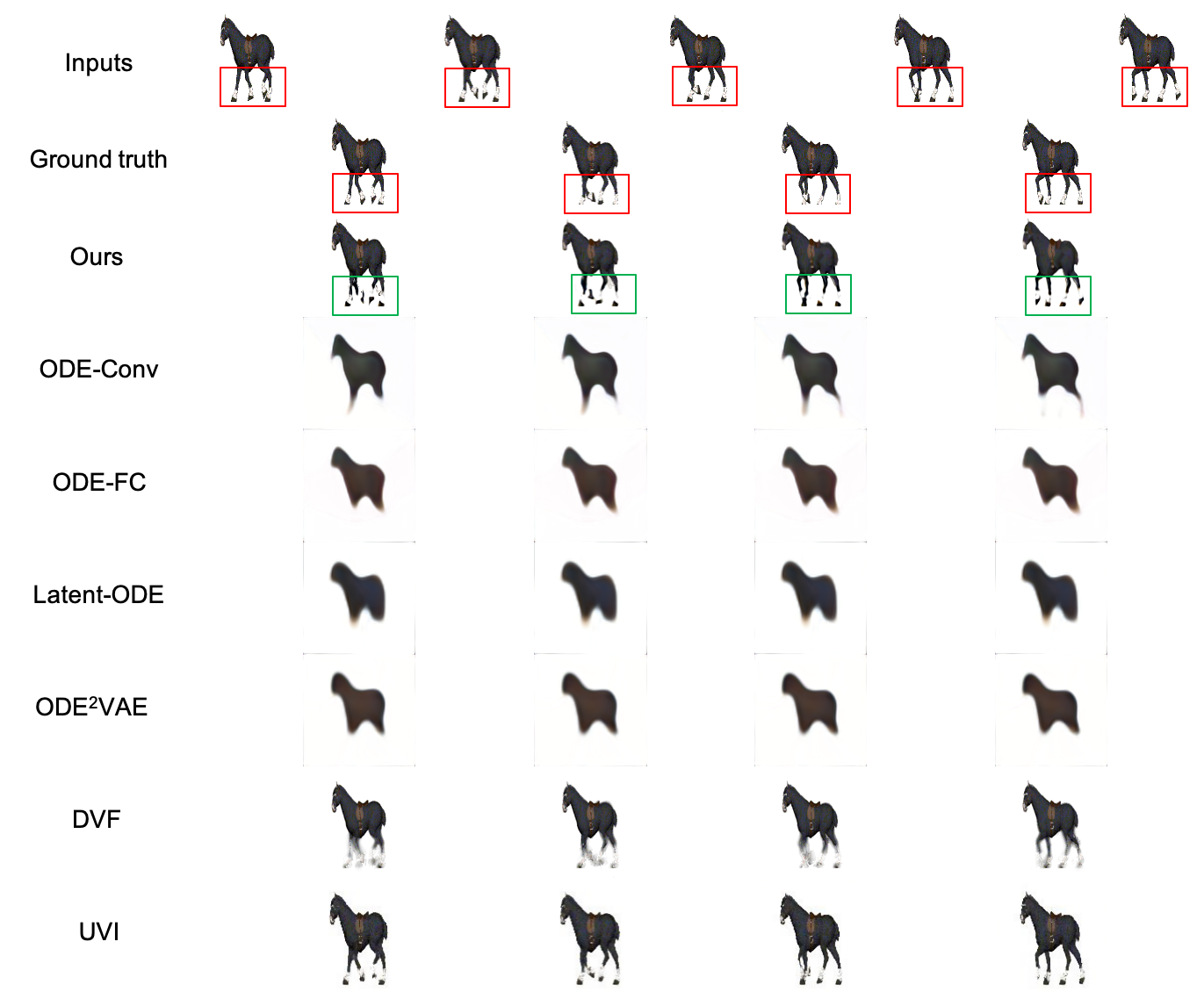}}
    \vspace{-0.1cm}
    \caption{Qualitative comparisons with interpolation baselines on the Moving GIF dataset.}
    \label{fig:base_interp_mgif}
    \vspace{-0.5cm}
\end{figure*}

\begin{figure*}[h]
    \centering
    \centerline{\includegraphics[width=0.7\linewidth]{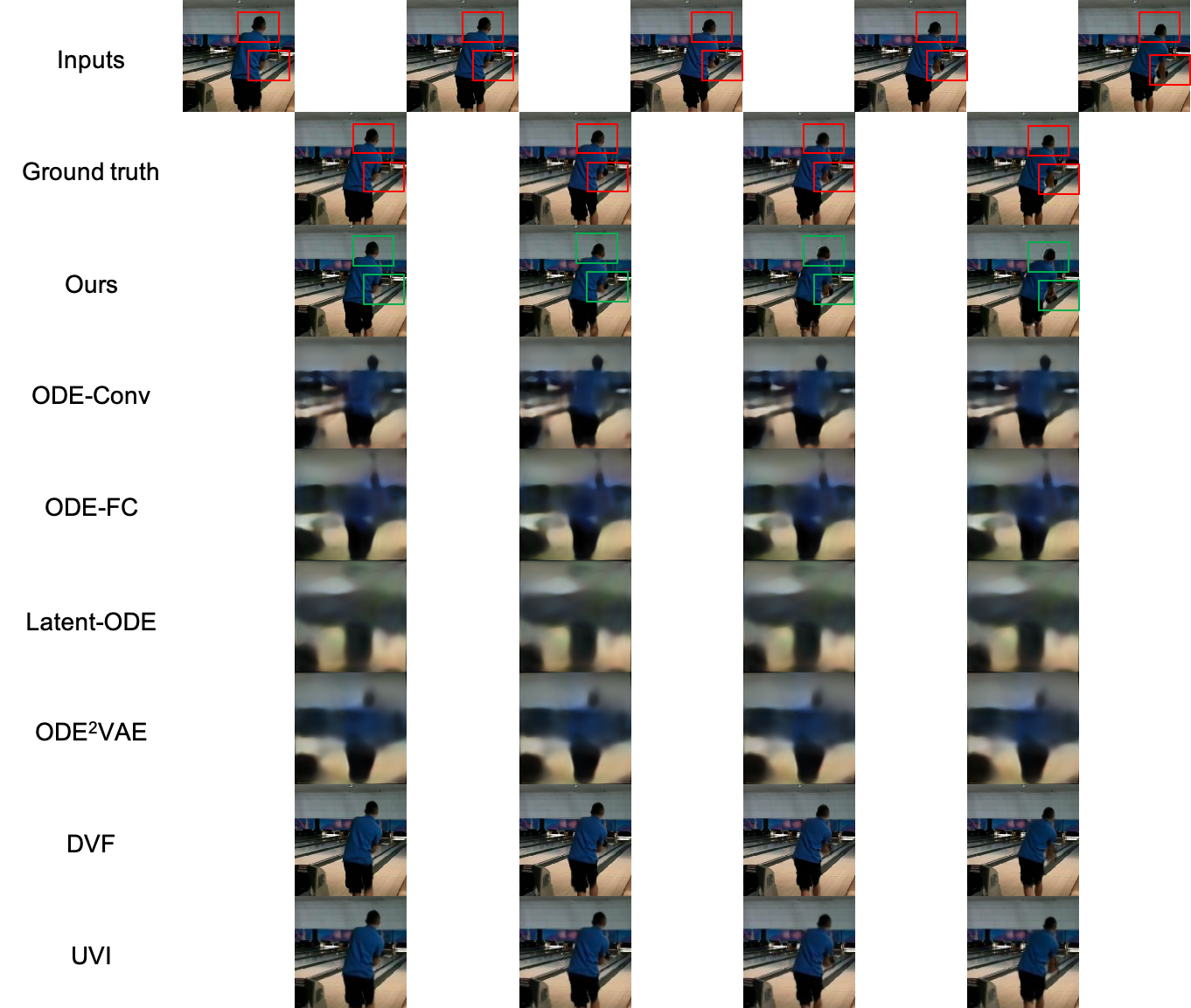}}
    \vspace{-0.1cm}
    \caption{Qualitative comparisons with interpolation baselines on the Penn Action dataset.}
    \label{fig:base_interp_penn}
    \vspace{-0.3cm}
\end{figure*}

\noindent\textbf{Video Interpolation.}
Figures~\ref{fig:base_interp_kth}--\ref{fig:base_interp_penn} illustrate video interpolation, where we generate 4 frames in-between the given 5 input frames. The locations with highly dynamic movements are marked with rectangles (\emph{red} for inputs and ground truth, \emph{green} for \model results).

\begin{figure*}[h]
    \centering
    \centerline{\includegraphics[width=0.7\linewidth]{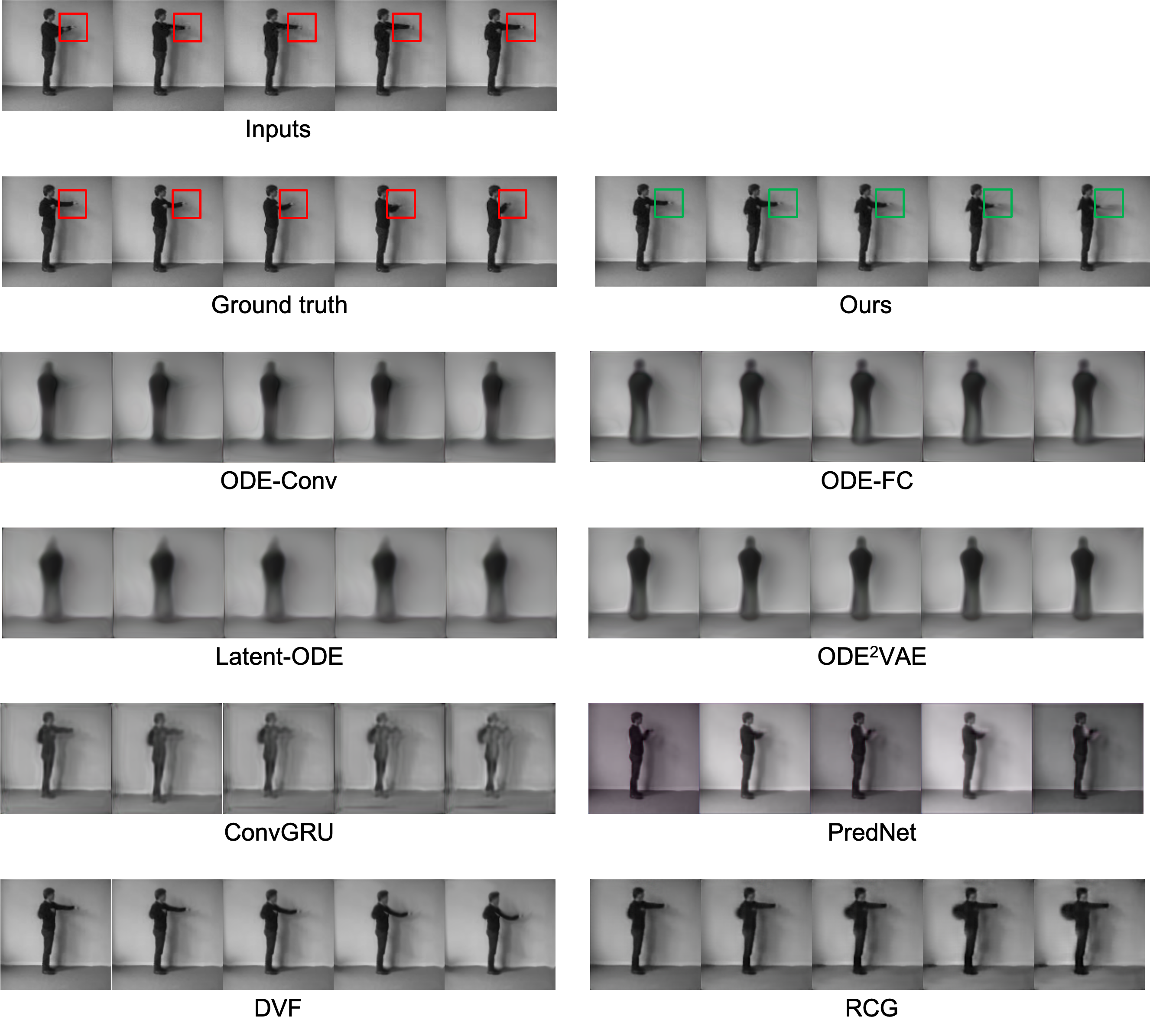}}
    \vspace{-0.2cm}
    \caption{Qualitative comparisons with extrapolation baselines on the KTH Action dataset.}
    \label{fig:base_extrap_kth}
    \vspace{-0.5cm}
\end{figure*}

\begin{figure*}[h]
    \centering
    \centerline{\includegraphics[width=0.7\textwidth]{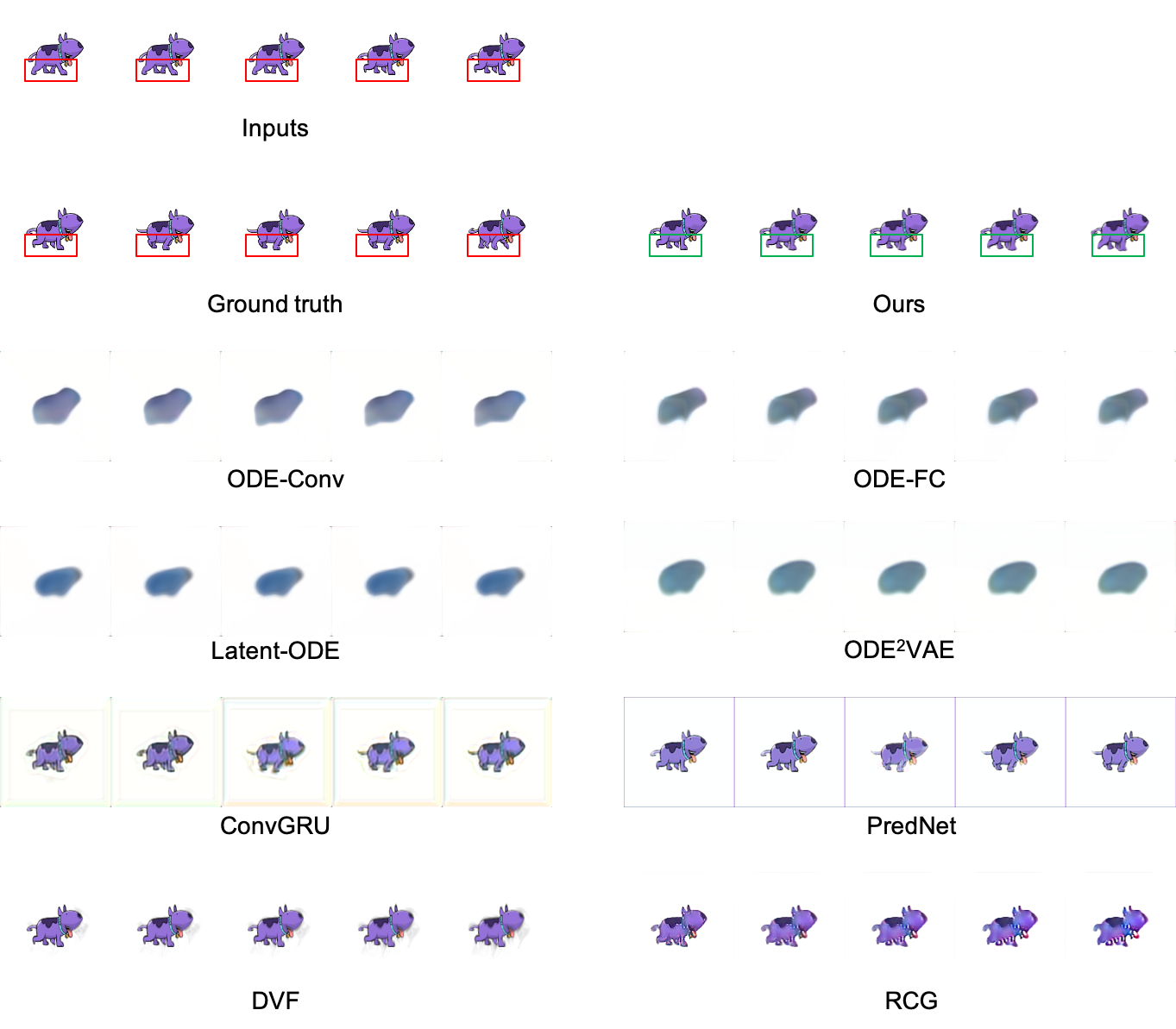}}
    \vspace{-0.4cm}
    \caption{Qualitative comparisons with extrapolation baselines on the Moving GIF dataset.}
    \label{fig:base_extrap_mgif}
    \vspace{-0.5cm}
\end{figure*}

\begin{figure*}[h]
    \centering
    \centerline{\includegraphics[width=0.7\textwidth]{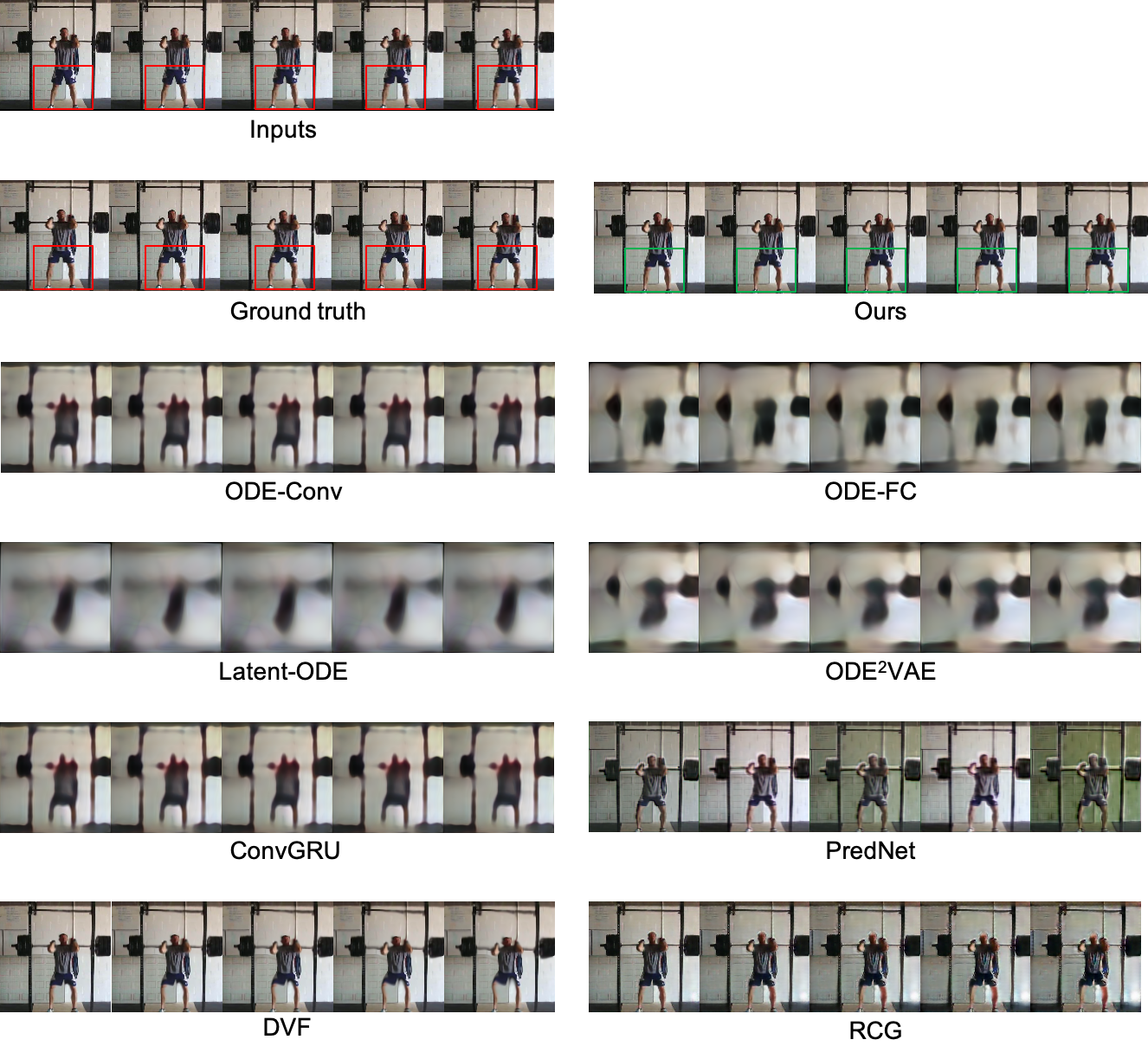}}
    \vspace{-0.4cm}
    \caption{Qualitative comparisons with extrapolation baselines on the Penn Action dataset.}
    \label{fig:base_extrap_penn}
    \vspace{-0.5cm}
\end{figure*}

\noindent\textbf{Video Extrapolation.}
Figures~\ref{fig:base_extrap_kth}--\ref{fig:base_extrap_penn} illustrate video extrapolation on KTH, Moving GIF, and Penn Action datasets, where we compare the 5 predicted future frames given 5 previous input frames. We mark the areas with highly dynamic movements in the same manner as video interpolation.

\begin{figure*}[h]
    \centering
    \centerline{\includegraphics[width=0.85\textwidth]{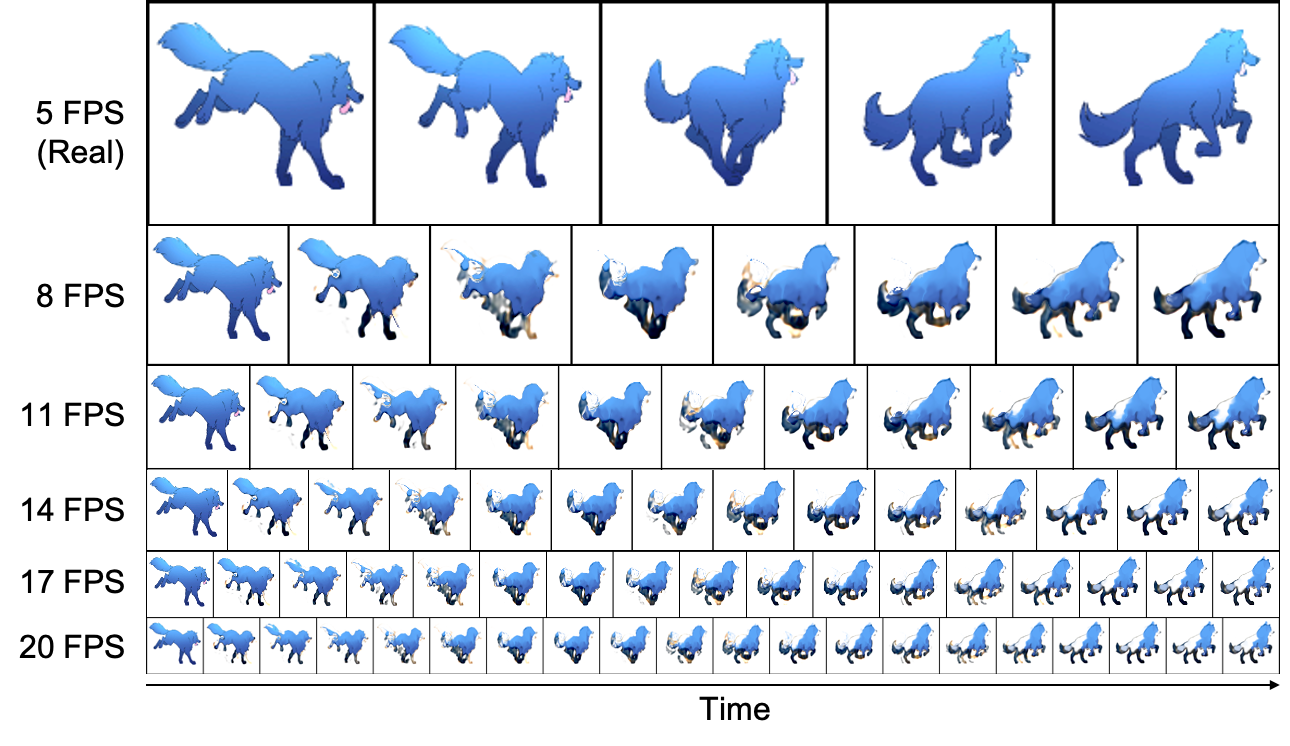}}
    \vspace{-0.5cm}
    \caption{Generated video frames in diverse time intervals based on a 5-FPS input video with the Moving GIF dataset. (\emph{Top row:} Input to \model. \emph{Remaining rows:} Videos generatedin various FPS between the start frame and the end frame.)}
    \vspace{-0.0cm}
    \label{fig:continuous_mgif}
\end{figure*}

\begin{figure*}[h]
    \centering
    \centerline{\includegraphics[width=0.85\textwidth]{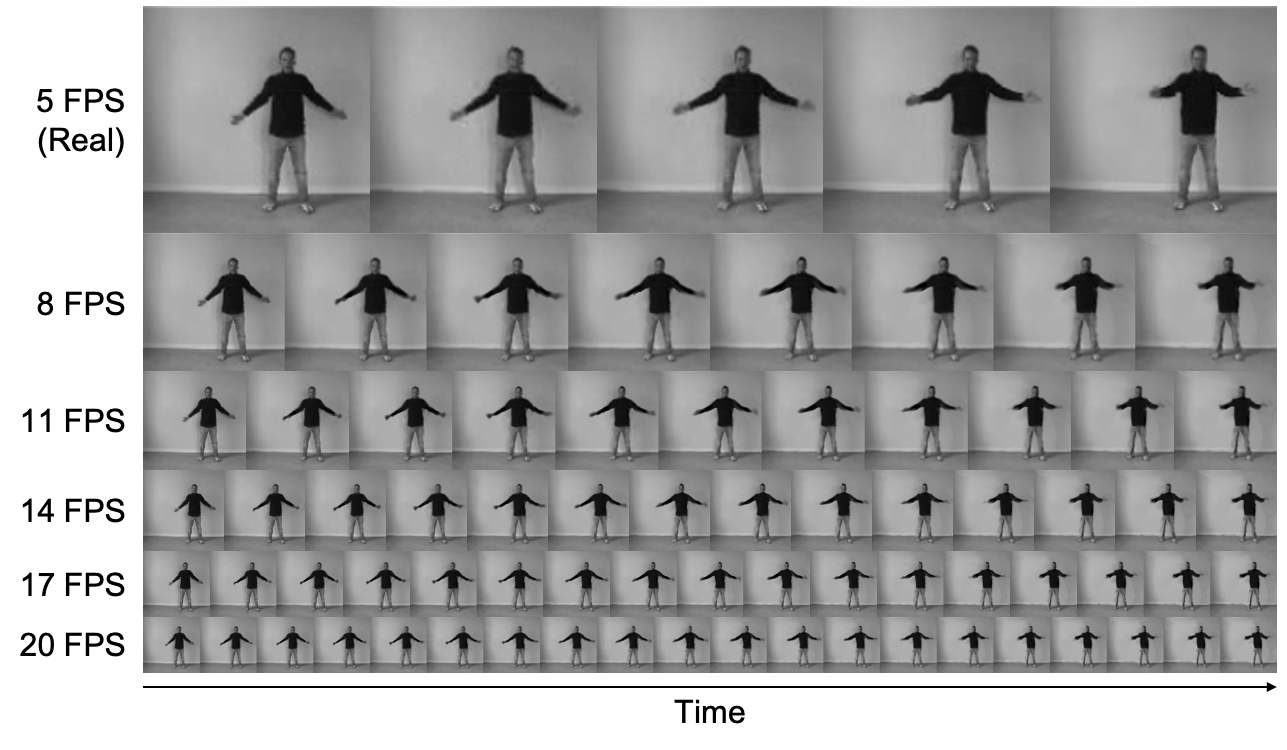}}
    \vspace{-0.5cm}
    \caption{Generated video frames in diverse time intervals based on a 5-FPS input video with the KTH Action dataset. (\emph{Top row:} Input to \model. \emph{Remaining rows:} Videos generated in various FPS between the start frame and the end frame.)}
    \label{fig:continuous_kth}
    \vspace{-0.5cm}
\end{figure*}

\noindent\textbf{Continuous Video Interpolation.}
In Figures~\ref{fig:continuous_mgif}--\ref{fig:continuous_kth}, given 5 frames as an input, we visualize videos generated in various FPS (\textit{e.g.,} 8, 11, 14, 17, and 20 FPS) between the first frame and the last frame.

\begin{figure*}[h]
    \centering
    \centerline{\includegraphics[width=0.85\textwidth]{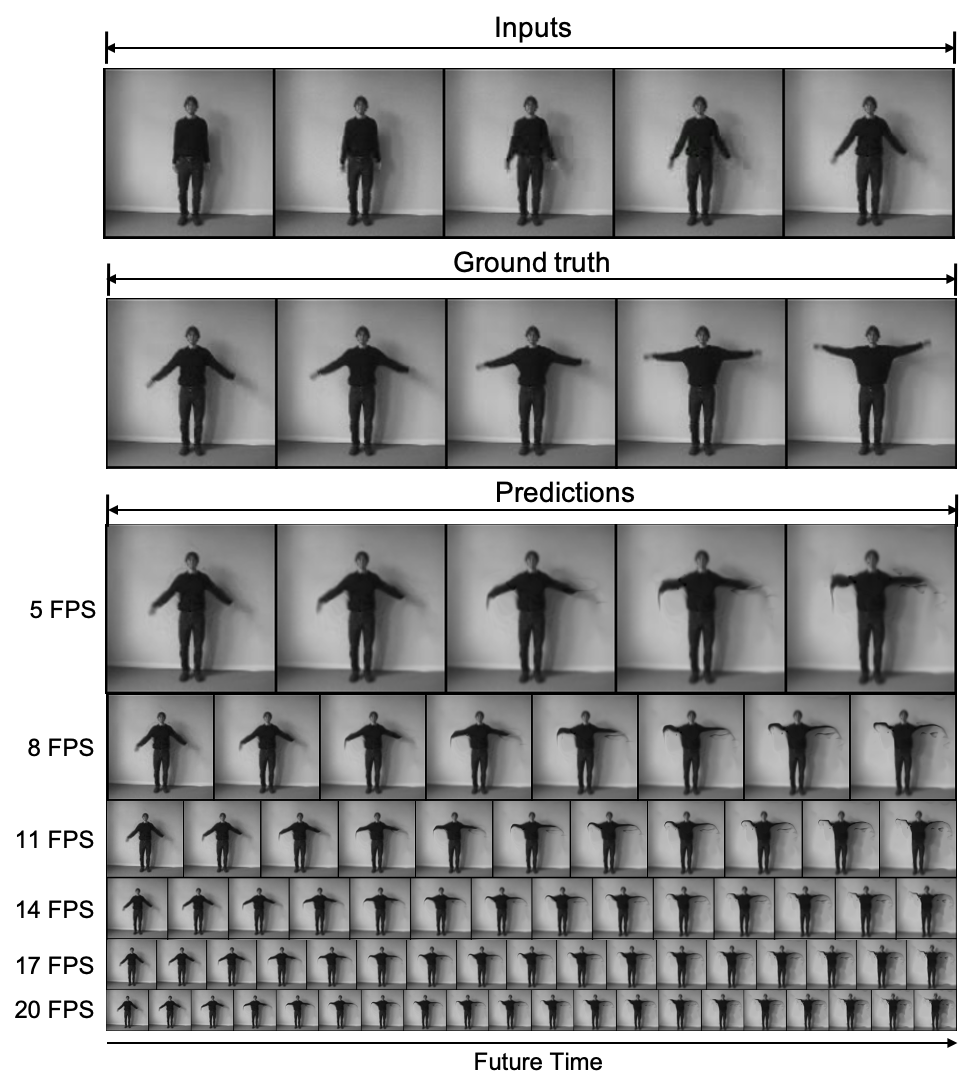}}
    \caption{Based on a 5-FPS input video of the KTH Action dataset, predicting and extrapolating video frames generated in diverse time intervals. (\emph{First row:} Input to \model. \emph{Second row:} Ground truth frames. \emph{Remaining rows:} Predicted or extrapolated frames in various FPS.)}
    \vspace{-0.5cm}
    \label{fig:continuous_extrap_kth}
\end{figure*}

\begin{figure*}[h]
    \centering
    \centerline{\includegraphics[width=0.85\textwidth]{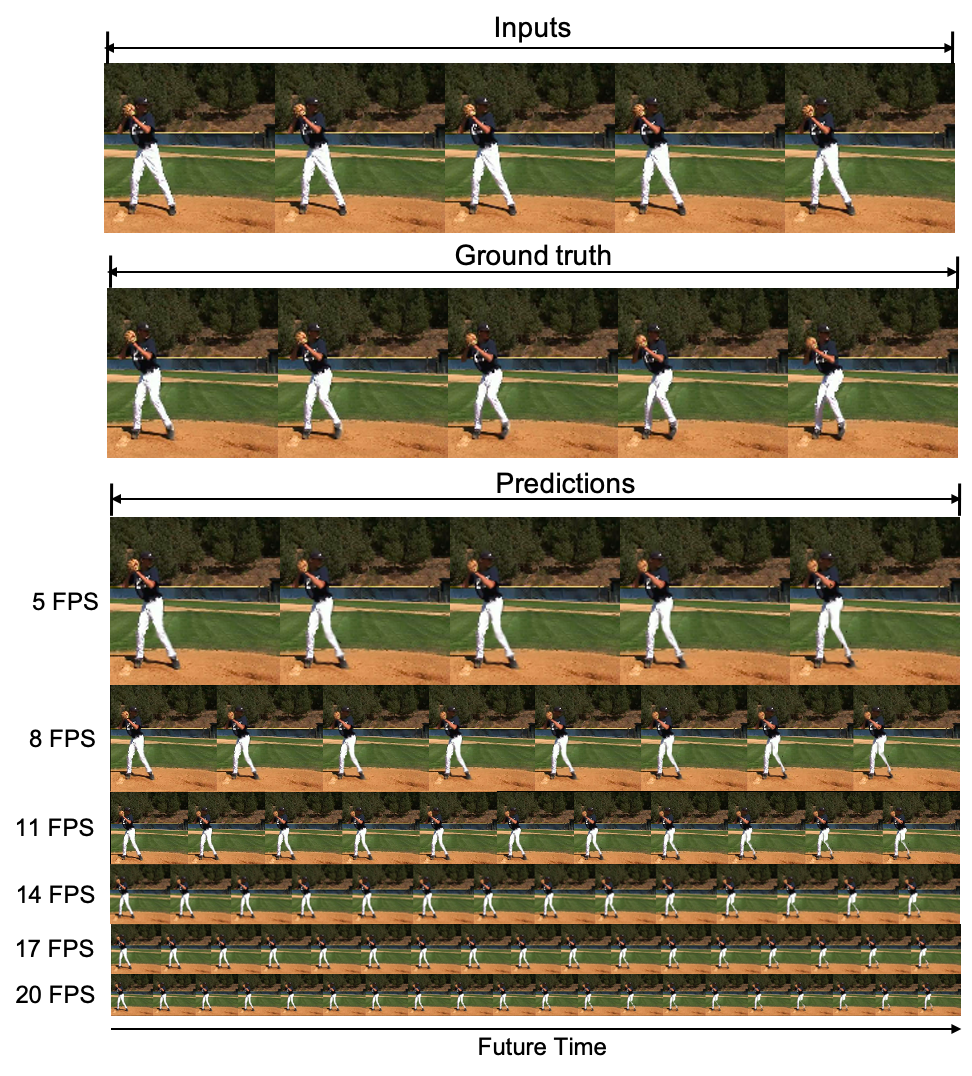}}
    \caption{Based on a 5-FPS input video of the Penn Action dataset, predicting and extrapolating video frames in diverse time intervals. (\emph{First row:} Input to \model. \emph{Second row:} Ground truth frames. \emph{Remaining rows:} Predicted or extrapolated frames in various FPS.)}
    \vspace{-0.5cm}
    \label{fig:continuous_extrap_penn}
\end{figure*}

\noindent\textbf{Continuous Video Extrapolation.}
In Figures~\ref{fig:continuous_extrap_kth}--\ref{fig:continuous_extrap_penn}, we generate future frames in various FPS (\textit{e.g.,} 8, 11, 14, 17, and 20 FPS)  after the last input frame.

\begin{figure*}
    \centering
    \includegraphics[width=\linewidth]{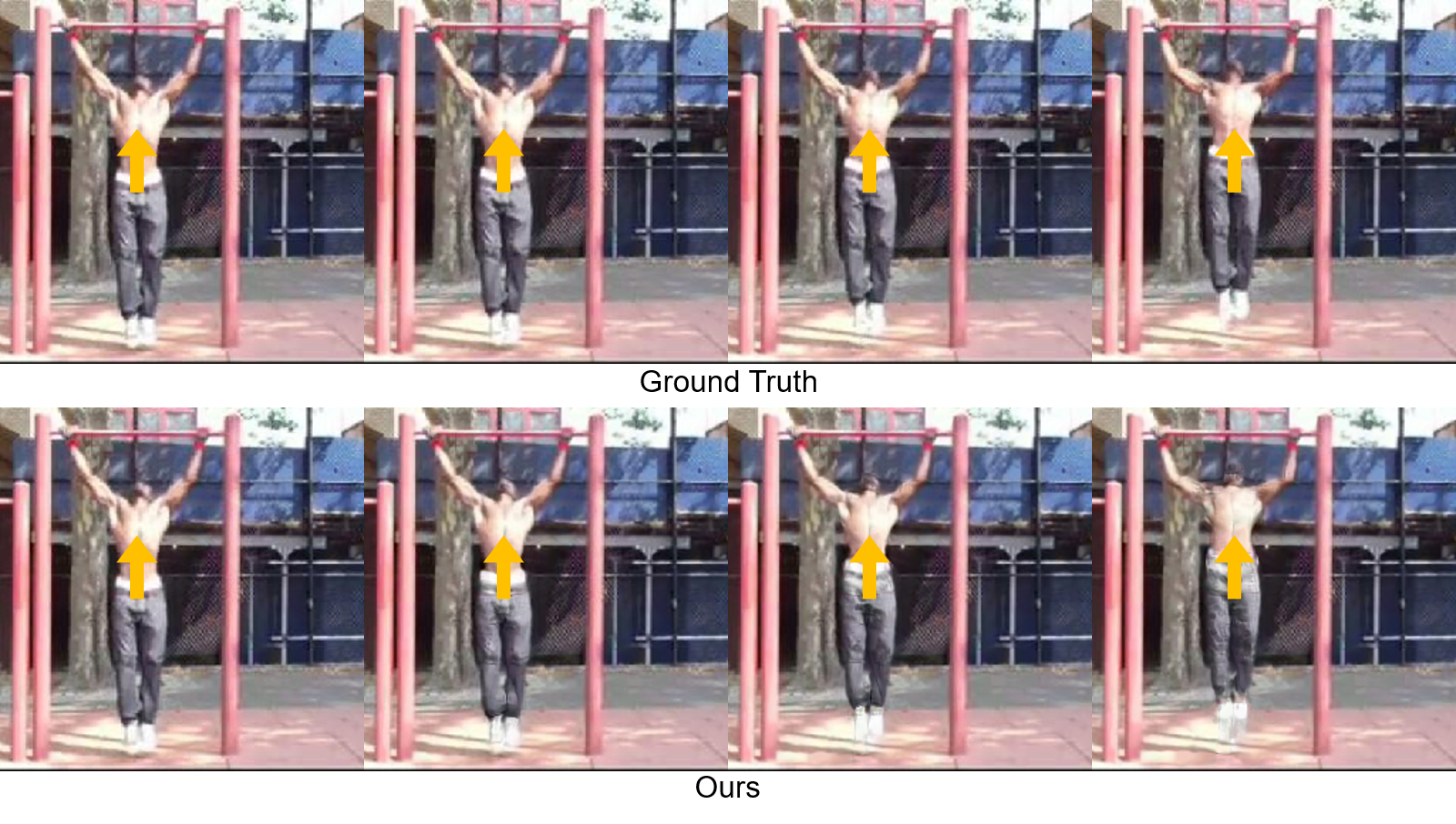}
    \includegraphics[width=\linewidth]{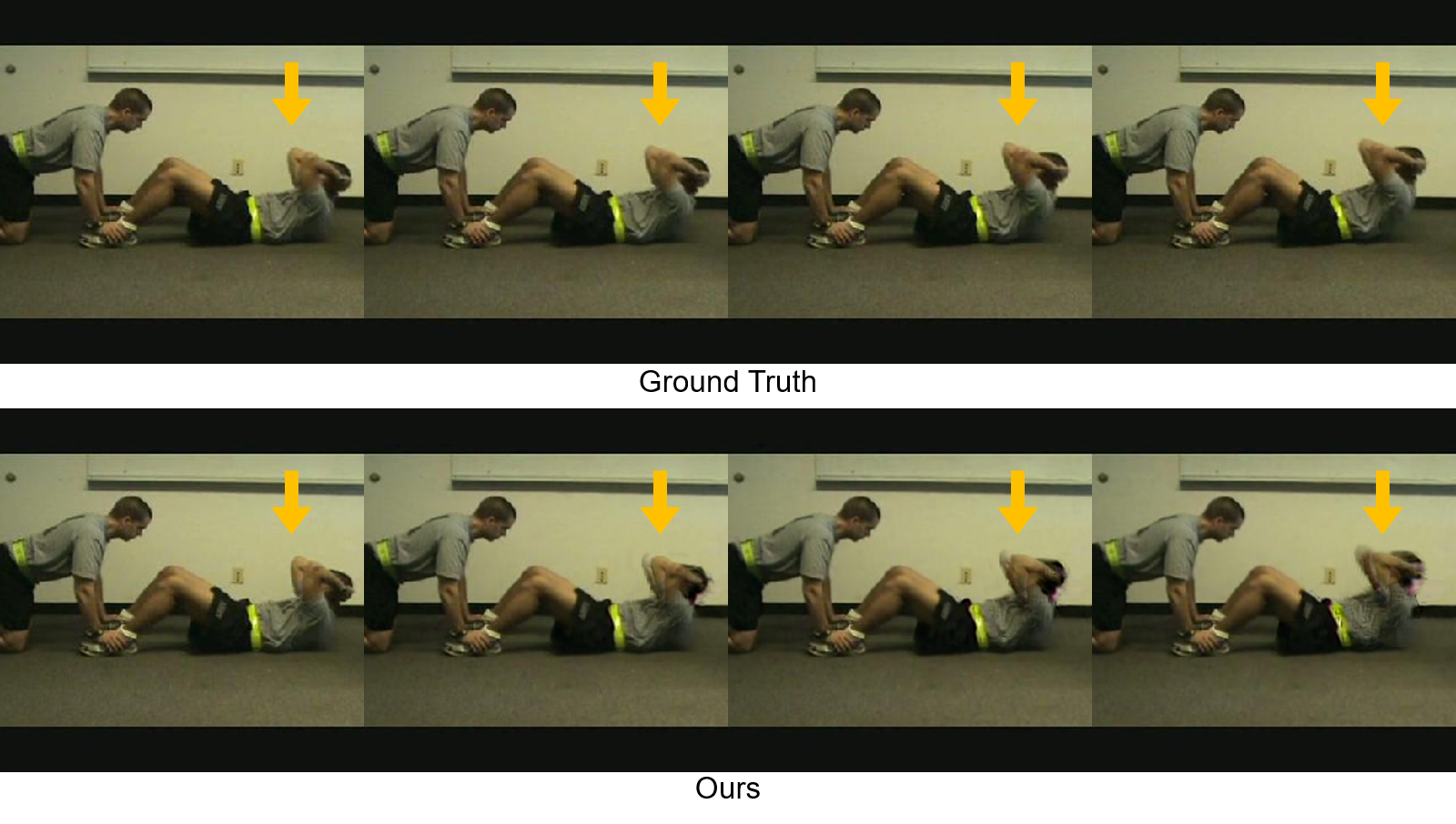}
    \caption{$256 \times 256$ resolution video interpolation results with Penn Action dataset.}
    \label{fig:HR_figure_interp}
\end{figure*}

\begin{figure*}
    \centering
    \includegraphics[width=\linewidth]{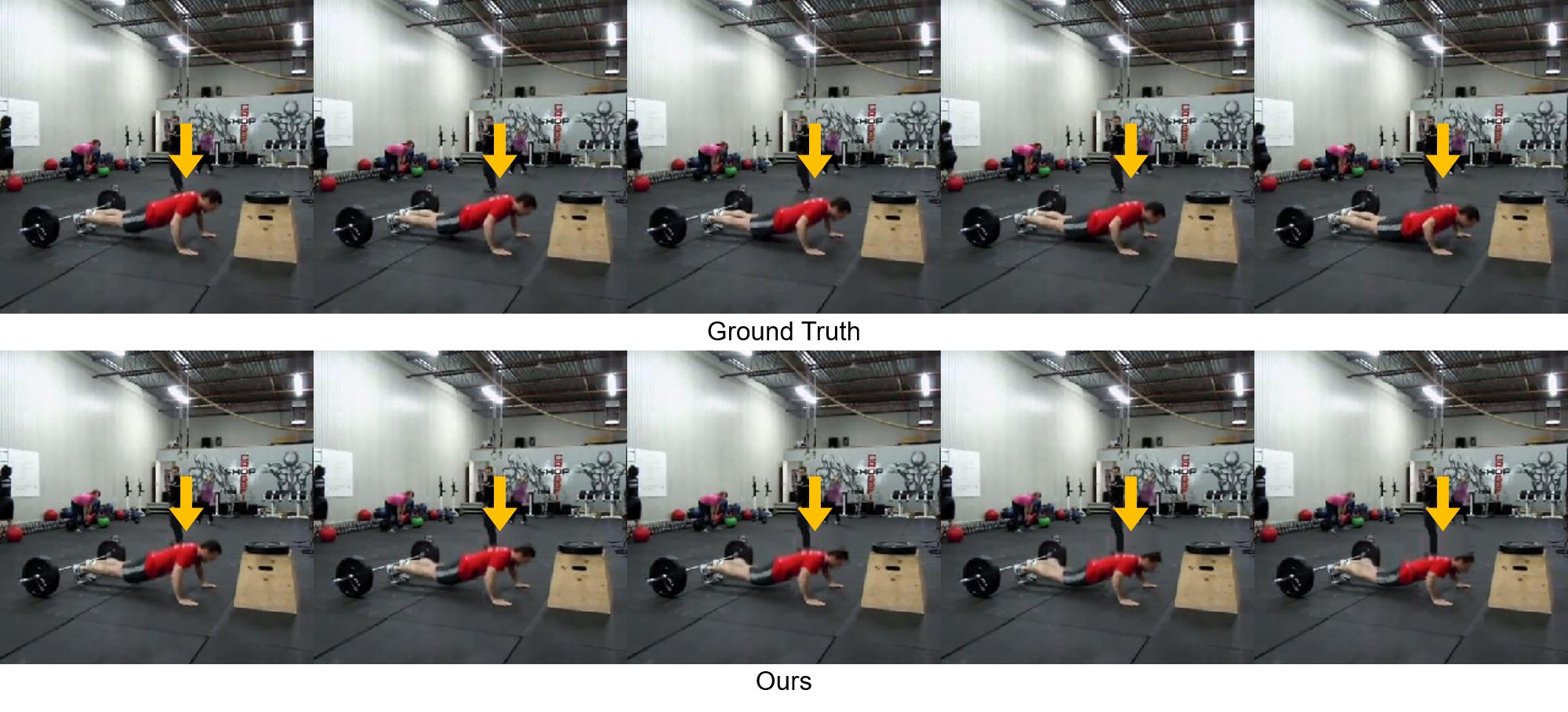}
    \includegraphics[width=\linewidth]{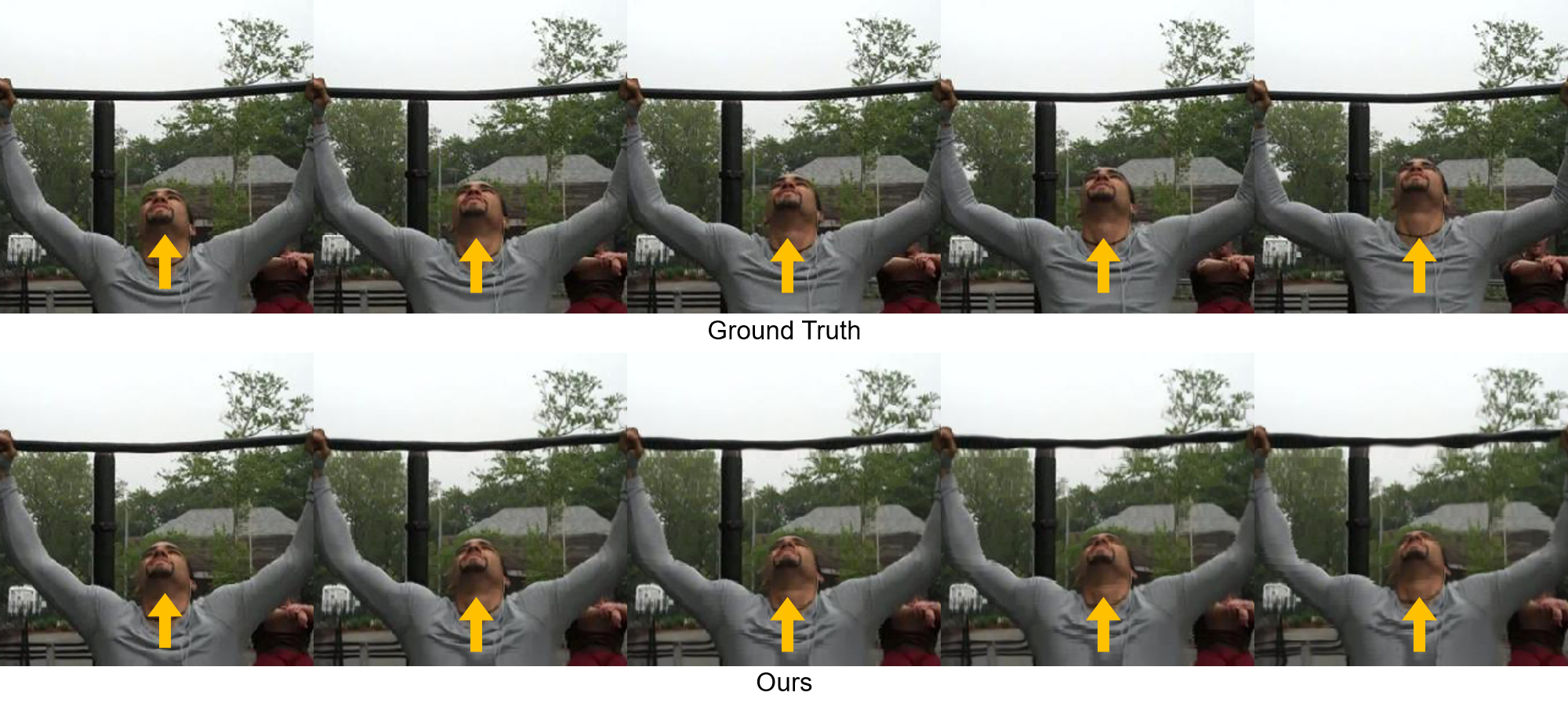}
    \caption{$256 \times 256$ resolution video extrapolation results with Penn Action dataset.}
    \label{fig:HR_figure_extrap}
\end{figure*}

\noindent\textbf{High-resolution video generation.}
As shown in Figure~\ref{fig:HR_figure_interp}--\ref{fig:HR_figure_extrap}, we further demonstrate that \model works well for higher resolution (256$\times$256) videos.
Moving parts and their directions are marked with \textit{yellow} arrows.

\begin{table*}[t!]
\begin{center}
\begin{tabular}{cccc}
\toprule
\textbf{Part} & \textbf{Layer} & \textbf{Output Shape} & \textbf{Layer Information}\\
\midrule
Input Image & - & $(h, w, c)$ & - \\
\midrule
\multirowcell{3}{Conv-Encoder $E$}
& Downsample & $(h/2, w/2, 32)$ & Conv-(N32, K3$\times$3, S1, P1), BN, ReLU \\
& Downsample & $(h/2, w/2, 64)$ & Conv-(N32, K4$\times$4, S2, P1), BN, ReLU \\
& Downsample & $(h/4, w/4, 128)$ & Conv-(N32, K4$\times$4, S2, P1), BN, ReLU \\
\midrule
\multirowcell{6}{ODE-ConvGRU}
& ConvGRU & $(h/4, w/4, 128)$ & Conv-(N128, K3$\times$3, S1, P1) \\
\cmidrule(lr){2-4}
& ODE Func $f_\theta$ & $(h/4, w/4, 128)$ & Conv-(N128, K3$\times$3, S1, P1), Tanh \\
& ODE Func $f_\theta$ & $(h/4, w/4, 128)$ & Conv-(N128, K3$\times$3, S1, P1), Tanh \\
& ODE Func $f_\theta$ & $(h/4, w/4, 128)$ & Conv-(N128, K3$\times$3, S1, P1), Tanh \\
& ODE Func $f_\theta$ & $(h/4, w/4, 128)$ & Conv-(N128, K3$\times$3, S1, P1) \\
\bottomrule
\end{tabular}
\vspace{-0.1cm}
\caption{Encoder architecture of our model.}
\label{Encoder Arch}
\end{center}
\end{table*}

\begin{table*}[t!]
\begin{center}
\begin{tabular}{cccc}
\toprule
\textbf{Part} & \textbf{Layer} & \textbf{Output Shape} & \textbf{Layer Information}\\
\midrule
\multirowcell{4}{ODE Solver}
& ODE Func $f_\phi$ & $(h/4, w/4, 128)$ & Conv-(N128, K3$\times$3, S1, P1), Tanh \\
& ODE Func $f_\phi$ & $(h/4, w/4, 128)$ & Conv-(N128, K3$\times$3, S1, P1), Tanh \\
& ODE Func $f_\phi$ & $(h/4, w/4, 128)$ & Conv-(N128, K3$\times$3, S1, P1), Tanh \\
& ODE Func $f_\phi$ & $(h/4, w/4, 128)$ & Conv-(N128, K3$\times$3, S1, P1) \\
\midrule
\multirowcell{3}{Conv-Decoder $G$}
& Upsample & $(h/2, w/2, 128)$ & Up, Conv-(N128, K3$\times$3, S1, P1), BN, ReLU \\
& Upsample & $(h, w, 64)$ & Up, Conv-(N64, K3$\times$3, S1, P1), BN, ReLU \\
& Upsample & $(h, w, c+3)$ & Up, Conv-(N(c+3), K3$\times$3, S1, P1) \\
\bottomrule
\end{tabular}
\vspace{-0.1cm}
\caption{Decoder architecture of our model.}
\label{Decoder Arch}
\end{center}
\end{table*}

\begin{table*}[t!]
\begin{center}
\begin{tabular}{cccc}
\toprule
\textbf{Part} & \textbf{Layer} & \textbf{Output Shape} & \textbf{Layer Information}\\
\midrule
Input Image & - & $(h, w, c)$ & - \\
\midrule
\multirowcell{5}{Discriminator $D_\text{img}$}
& Downsample & $(h/2, w/2, 64)$ & Conv-(N64, K4$\times$4, S2, P1), LeakyReLU \\
& Downsample & $(h/4, w/4, 128)$ & Conv-(N128, K4$\times$4, S2, P1), LeakyReLU \\
& Downsample & $(h/8, w/8, 256)$ & Conv-(N256, K4$\times$4, S2, P1), LeakyReLU \\
& Downsample & $(h/8, w/8, 512)$ & Conv-(N512, K4$\times$4, S1, P2), LeakyReLU \\
& Downsample & $(h/8, w/8, 64)$ & Conv-(N64, K4$\times$4, S1, P2) \\
\midrule
Input Sequence & - & $(h, w, c \times t)$ & - \\
\midrule
\multirowcell{5}{Discriminator $D_\text{seq}$}
& Downsample & $(h/2, w/2, 64)$ & Conv-(N64, K4$\times$4, S2, P1), LeakyReLU \\
& Downsample & $(h/4, w/4, 128)$ & Conv-(N128, K4$\times$4, S2, P1), LeakyReLU \\
& Downsample & $(h/8, w/8, 256)$ & Conv-(N256, K4$\times$4, S2, P1), LeakyReLU \\
& Downsample & $(h/8, w/8, 512)$ & Conv-(N512, K4$\times$4, S1, P2), LeakyReLU \\
& Downsample & $(h/8, w/8, 64)$ & Conv-(N64, K4$\times$4, S1, P2) \\
\bottomrule
\end{tabular}
\vspace{-0.1cm}
\caption{Discriminator architecture of our model.}
\label{Discriminator Arch}
\end{center}
\end{table*}

\begin{table*}[t!]
\begin{center}
\begin{tabular}{p{0.1\textwidth} p{0.06\textwidth}<{\centering} p{0.06\textwidth}<{\centering} p{0.06\textwidth}<{\centering} p{0.06\textwidth}<{\centering} p{0.06\textwidth}<{\centering} p{0.06\textwidth}<{\centering} p{0.06\textwidth}<{\centering} p{0.06\textwidth}<{\centering} p{0.06\textwidth}<{\centering}}
\toprule
\multicolumn{1}{c}{\multirow{2}{*}{\textbf{Model}}}
& \multicolumn{3}{c}{\textbf{KTH Action}} & 
\multicolumn{3}{c}{\textbf{Moving GIF}} & 
\multicolumn{3}{c}{\textbf{Penn Action}} \\
\cmidrule(lr){2-4} \cmidrule(lr){5-7} \cmidrule(lr){8-10}
 & SSIM$_{\uparrow}$ &  LPIPS$_{\downarrow}$ & PSNR$_{\uparrow}$ 
 & SSIM$_{\uparrow}$ & LPIPS$_{\downarrow}$ & PSNR$_{\uparrow}$
 & SSIM$_{\uparrow}$ & LPIPS$_{\downarrow}$ & PSNR$_{\uparrow}$\\
\midrule
SVG-FP
 & 0.848 &\ 0.156 &\ 26.34
 & 0.682 &\ 0.254 &\ 15.62
 & 0.815 &\ 0.099 &\ 21.53
\\
SVG-LP
 & 0.843 &\ 0.157 &\ 22.24
 & 0.710 &\ 0.233 &\ 15.95
 & 0.814 &\ 0.117 &\ 21.36
\\
SAVP
 & \textbf{0.885} &\ 0.089 &\ 27.40
 & 0.722 &\ 0.163 &\ 15.66
 & 0.822 &\ 0.074 &\ 22.23
\\
SRVP
 & 0.831 &\ 0.113 &\ 27.79
 & 0.667 &\ 0.263 &\ 16.23
 & 0.877 &\ 0.056 &\ 22.10
\\
\midrule
\texttt{\model}
 & 0.878 &\ \textbf{0.080} &\ \textbf{28.19}
 & \textbf{0.778} &\ \textbf{0.156} &\ \textbf{16.68}
 & \textbf{0.880} &\ \textbf{0.045} &\ \textbf{23.81}
\\
\bottomrule
\end{tabular}
\vspace{-0.1cm}
\caption{Additional video extrapolation comparisons. We report the baseline results by taking the average of three samples.}
\label{tab:comparison}
\end{center}
\end{table*}

\clearpage

\end{document}